\newcount\Comments  % 0 suppresses notes to selves in text
\documentclass[letterpaper,10pt]{scrartcl}

\usepackage[utf8]{inputenc}
\usepackage[english]{babel}
\usepackage{amsmath,amssymb,upref, bbm}
\usepackage{amsfonts}
\usepackage{amsthm,array,makecell}
\usepackage{color}
\usepackage{graphicx}
\usepackage{sidecap}
\usepackage{multirow}
\usepackage{booktabs}
\usepackage{fullpage}
\usepackage[title]{appendix}

\usepackage{tikz}
\usetikzlibrary{shapes}
\usetikzlibrary{positioning}
\usetikzlibrary{matrix}
\usetikzlibrary{patterns}
\usetikzlibrary{automata, positioning}

\usepackage[font=small]{caption}
\usepackage{subcaption}

\usepackage{longtable}
\usepackage{rotating}

\usepackage{algorithm}
\usepackage{algcompatible}
\usepackage[noend]{algpseudocode}
\extrafloats{100}
\usepackage{lineno}
\usepackage[hidelinks]{hyperref}
\usepackage{subcaption}

\usepackage{listings}

\graphicspath{{figures/}}
\DeclareGraphicsExtensions{.pdf,.eps,.png,.jpg,.jpeg}

\newcommand\Algphase[1]{%
\vspace*{-.2\baselineskip}\Statex\hspace*{\dimexpr-\algorithmicindent-2pt\relax}\rule{\textwidth}{0.4pt}%
\Statex\hspace*{-\algorithmicindent}\textbf{#1}%
\vspace*{-.7\baselineskip}\Statex\hspace*{\dimexpr-\algorithmicindent-2pt\relax}\rule{\textwidth}{0.4pt}%
}

\newcommand{\R}{\mathbb{R}}

\newcommand{\bfx}{\boldsymbol x}
\newcommand{\bfX}{\boldsymbol X}

\newcommand{\bfb}{\boldsymbol b}

\newcommand{\bfr}{\boldsymbol r}
\newcommand{\bfu}{\boldsymbol u}

\newcommand{\bfz}{\boldsymbol z}

\newcommand{\bfC}{\boldsymbol C}

\newcommand{\bfK}{\boldsymbol K}
\newcommand{\bfL}{\boldsymbol L}

\newcommand{\bfU}{\boldsymbol U}
\newcommand{\bfV}{\boldsymbol V}

\newcommand{\bfQ}{\boldsymbol Q}

\newcommand{\bfW}{\boldsymbol W}

\newcommand{\tbfC}{\tilde{\bfC}}
\newcommand{\tbfU}{\tilde{\bfU}}
\newcommand{\tbfx}{\tilde{\bfx}}

\newcommand{\tbfL}{\tilde{\bfL}}

\newcommand{\bfTcal}{\boldsymbol{\mathcal{T}}}
\newcommand{\bfXcal}{\boldsymbol{\mathcal{X}}}
\newcommand{\bfYcal}{\boldsymbol{\mathcal{Y}}}
\newcommand{\bfZcal}{\boldsymbol{\mathcal{Z}}}
\newcommand{\tbfXcal}{\tilde{\bfXcal}}
\newcommand{\hatx}{\hat{x}}

\newcommand{\nSites}{m}

\newcommand{\timeEnd}{T}

\newcommand{\timeEndObs}{T_{\text{obs}}}

\newcommand{\timeEndSim}{T_{\text{sim}}}

\newcommand{\nTimes}{n}

\newcommand{\nTimesObs}{n_{\text{obs}}}

\newcommand{\nTimesSim}{n_{\text{sim}}}

\newcommand{\fsx}{\hat{\bfx}}

\newcommand{\ty}{\tilde{y}}

\newcommand{\qInEnc}{q^{\text{enc}}_{\text{in}}}

\newcommand{\qOut}{q_{\text{out}}}

\newcommand{\qInDec}{q^{\text{dec}}_{\text{in}}}

\newcommand{\nClusters}{n_{\text{clusters}}}

\newcommand{\dmodel}{d_{\text{model}}}

\newcommand{\dff}{d_{\text{ff}}}

\newcommand{\nHeads}{n_{\text{head}}}

\newcommand{\nEnc}{n_{\text{enc}}}

\newcommand{\nDec}{n_{\text{dec}}}

\newcommand{\nMarkov}{n_{\text{Markov}}}

\newcommand{\qmax}{q_{\text{max}}}

\newcommand{\MarkovLag}{p}

\newcommand{\kibitz}[2]{\ifnum\Comments=1\textcolor{#1}{#2}\fi}

\newenvironment{keywords}%
   {\begin{trivlist}\item[]{\bfseries\sffamily Keywords:}\ }% oder "Keywords:"
   {\end{trivlist}}

\numberwithin{equation}{section}

\title{GenFormer: A Deep-Learning-Based Approach for Generating Multivariate Stochastic Processes}

\author{Haoran Zhao\thanks{hz289@cornell.edu}, Wayne Isaac Tan Uy\thanks{wtu4@cornell.edu}}
\date{\today}

\begin{document}

\maketitle

\begin{abstract}
Stochastic generators are essential to produce synthetic realizations that preserve target statistical properties. We propose GenFormer, a stochastic generator for spatio-temporal multivariate stochastic processes. It is constructed using a Transformer-based deep learning model that learns a mapping between a Markov state sequence and time series values. The synthetic data generated by the GenFormer model preserves the target marginal distributions and approximately captures other desired statistical properties even in challenging applications involving a large number of spatial locations and a long simulation horizon. The GenFormer model is applied to simulate synthetic wind speed data at various stations in Florida to calculate exceedance probabilities for risk management. 
\end{abstract}

\begin{keywords}
stochastic generator, multivariate stochastic processes, Transformer-based deep learning model, Markov processes, time series forecasting
\end{keywords}

\section{Introduction}

        Stochastic generators are tools to produce synthetic data which preserves desired statistical properties. They are crucial in situations wherein the number of available records is inadequate while a large amount of data is required. This is especially the case in reliability analysis and risk management. For example, performance-based engineering requires synthetic data that characterizes excitations from natural hazards to provide accurate reliability estimates of building systems \cite{pbe_wind, pbe_earthquake}. Robust risk assessments of parametric insurance products necessitate the generation of supplemental synthetic loss events for various perils such as hurricane and excess rainfall \cite{parametric_insurance, parametric_rainfall_insurance}. The underlying data in the aforementioned examples, e.g., wind pressure fields and precipitation time series, are typically spatio-temporal in nature and can be represented as multivariate stochastic processes.

        While developing stochastic generators for multivariate Gaussian processes is a well-established research area, constructing models which can be applied to generate time series with consistent non-Gaussian features remains to be a challenge. One class of methods constitutes direct extensions of those used for Gaussian processes which target certain statistical properties beyond the second moment. The third-order spectral representation method introduced in \cite{SRM_3_model} appends additional terms to the traditional spectral representation which is typically employed to simulate Gaussian processes. This enables it to capture the third-moment properties. The polynomial chaos \cite{pc_model} and translation processes \cite{translation_paper_mix} are nonlinear transformations of Gaussian processes. The former utilizes truncated Hermite polynomials while the latter applies a transformation based on marginal distributions to approximately match the finite dimensional distributions and exactly match the marginal distributions, respectively. Recently, a mix of the above two models is proposed in \cite{pct_model}. In addition, stochastic generators based on the Markov assumption have been formulated to produce synthetic time series approximating finite dimensional distributions. These incorporate a variety of techniques such as the resampling procedure \cite{precipitation_paper}, $K$-nearest neighbor algorithm \cite{markov_knn}, and copula models \cite{haoran_thesis}. A prevalent challenge in many of the aforementioned models is the curse of dimensionality in which increasing the number of variates, i.e. spatial locations, and the simulation time horizon leads to substantial computational demands or significant decline in model performance of approximating target statistical properties \cite{pc_dimension_issue}. 

        Deep learning models, a subset of machine learning algorithms, have gained prominence in recent years due to their capabilities in solving large-scale problems. These models excel in feature extraction and pattern recognition involving medium to large datasets, making them well-suited for complex tasks such as image and speech recognition \cite{deep_learning_image_recognition, deep_learning_speech_recognition}, natural language processing \cite{cnn_rnn_nlp, attention_paper}, and time series forecasting \cite{rnn_forecasting, trans_forecasting}. Time series forecasting is concerned with predicting future time series values based on the historical records. Recent deep forecasting models, particularly those based on the Transformer architecture, have achieved great progress in predicting multivariate processes with a large number of locations over a long time horizon \cite{informer_paper, autoformer_paper}. Due to the attention mechanism, Transformers are capable of modeling long-term dependencies and patterns in sequential data, leading to markedly-improved prediction accuracy.
        
        Inspired by the use of deep learning models in forecasting applications, we propose GenFormer, a deep-learning-based stochastic generator for stationary and ergodic multivariate processes with continuous marginal distributions which aims to tackle the challenges of simulation in high dimensions. GenFormer is constructed under the Markov assumption and can be regarded as an extension of \cite{precipitation_paper}. It is composed of two models. The first is a univariate discrete-time Markov process in which each type of spatial variation across locations is represented by a Markov state. The second is a Transformer-based deep learning model which establishes a mapping from the Markov states to the values of time series. The generation of synthetic realizations of multivariate processes based on the GenFormer begins with the simulation of a synthetic univariate Markov sequence which subsequently serves as input for the deep-learning-based mapping. In practice, the synthetic data generated by the deep learning model may not preserve essential statistical properties such as the spatial correlation and marginal distributions. As such, the GenFormer model incorporates a model post-processing step involving a transformation of the resulting samples based on the Cholesky decomposition as well as a sample reshuffling technique to correct for key statistical properties. The final synthetic data produced by GenFormer is able to exactly match the marginal distributions and approximately match  other statistical properties, including higher-order moments or probabilities of quantities of interest. Our numerical examples involving numerous spatial locations and simulation over a long time horizon demonstrate that synthetic realizations produced by GenFormer can be reliably utilized in downstream applications due to the superior performance of deep learning models for complex and high-dimensional tasks. 
        
        The paper is organized as follows. Section~\ref{sec:prelim} outlines preliminaries of the GenFormer model. We review the stochastic generator proposed in \cite{precipitation_paper} designed for precipitation data as well as deep learning models used for time series forecasting. The construction and simulation algorithm of the GenFormer model are presented in Section~\ref{sec:newmodel}. The performance of GenFormer on approximating the target statistical properties of interest is examined in Section~\ref{sec:numerical_examples} via numerical examples involving a synthetic dataset generated from stochastic differential equations and a real dataset of wind speeds measured at stations in Florida. Concluding remarks are offered in Section~\ref{sec:conclusion}.

\section{Preliminaries}
\label{sec:prelim}

We present in Section~\ref{subsec:prelim:stochastic_generator} a stochastic generator tailored to precipitation data. Existing Transformer-based deep learning models used for time series forecasting are then summarized in Section~\ref{subsec:prelim:deep_learning_model}. These provide the foundation for the proposed stochastic generator in this work.

\subsection{Stochastic generator for $\nSites$-station rainfall processes}
\label{subsec:prelim:stochastic_generator}

Let $\bfX(t) = [X_1(t), \dots, X_{\nSites}(t)]^T, t \in [0, \timeEnd]$, be a $\nSites$-variate stationary and ergodic stochastic process with continuous marginal distributions where $X_i(t)$ is a univariate stochastic process modeling the temporal evolution of a quantity of interest at spatial location $i$, $i=1, \dots, \nSites$. In practice, $\bfX(t)$ is measured at $\nTimes$ discrete time stamps $t_1, \ldots, t_{\nTimes}$, $0 = t_1 < \cdots < t_{\nTimes} = \timeEnd$, that evenly partition the time interval $[0,\timeEnd]$. Stochastic generators, fitted using the observed realizations $\bfx(t) = [x_1(t),\dots,x_{\nSites}(t)]^T$ of $\bfX(t)$, produce additional synthetic realizations that preserve the statistical properties of $\bfX(t)$. These synthetic realizations can then be employed in downstream applications such as estimating exceedance probabilities of quantities of interest derived from $\bfX(t)$.

The stochastic generator proposed in \cite{precipitation_paper} is of interest in this subsection. It is designed for multi-station precipitation data, i.e., $X_i(t)$ corresponds to a rainfall process measured at station $i$ with $x_i(t)$ denoting the observed data. The construction of the stochastic generator involves three steps. First, a univariate Markov sequence $Y_1, \ldots, Y_{\nTimes}$ that relates to the spatial rainfall pattern across stations is constructed from $\bfX(t_1), \ldots, \bfX(t_{\nTimes})$. Second, a synthetic realization $\ty_1, \ldots, \ty_{\nTimes}$ of the Markov state sequence is generated which guides the simulation of the preliminary synthetic realization $\tbfx(t_1), \ldots, \tbfx(t_{\nTimes})$ of the rainfall sequence via resampling. Third, the final synthetic sequence $\fsx(t_1), \ldots, \fsx(t_{\nTimes})$ is obtained using the reshuffling technique according to the ranks of the realization in the previous step. We discuss these steps further in the subsections below.

% in a semi-parametric manner
%New realizations of the Markov state sequences $\ty_1, \ldots, \ty_{\nTimes}$ can then be generated iteratively based on the estimated Markov order and the corresponding transition matrix. Second, the preliminary synthetic realizations of the rainfall sequence $\tbfx(t_1), \ldots, \tbfx(t_{\nTimes})$ are generated by applying the resampling mapping to the rainy sequences within $\ty_1, \ldots, \ty_{\nTimes}$. Third, the final synthetic sequence $\fsx(t_1), \ldots, \fsx(t_{\nTimes})$ are obtained based on the reshuffling technique, which arranges the newly-simulated precipitation values from parametric distribution according to the ranks of $\tbfx(t_1), \ldots, \tbfx(t_{\nTimes})$ per station. We discuss these steps further in the sections below.

\subsubsection{Univariate Markov sequences for $\nSites$-variate stochastic processes}
\label{subsubsec:prelim:markov_sequence}

The approach proceeds by fitting a univariate Markov process $Y_1, Y_2, \ldots, Y_{\nTimes}$ corresponding to $\bfX(t_j)$ at time stamps $t_j, j = 1, \ldots, \nTimes$.  The stochastic process $Y_1, Y_2, \ldots, Y_{\nTimes}$ is a $\MarkovLag^{\text{th}}$-order discrete-time Markov process if the conditional random variables $(Y_j| Y_{j-1},\ldots,Y_{j-\MarkovLag})$ and $(Y_{j-\MarkovLag-1},\ldots,Y_1| Y_{j-1},\ldots,Y_{j-\MarkovLag} )$ are independent for $j > \MarkovLag+1$ \cite{markov_property}. Under the assumption that $Y_j$ is discrete-valued, the above definition readily implies the well-known Markov property
\begin{equation} 
P(Y_j = y_j|Y_{j-1} = y_{j-1},\ldots,Y_1 = y_1) = P(Y_j = y_j|Y_{j-1} = y_{j-1},\ldots,Y_{j-\MarkovLag} = y_{j-\MarkovLag}),~~j \ge \MarkovLag+1\label{implication of Markov}.
\end{equation}
The state $Y_j$ at time stamp $t = t_j$ thus depends only upon the states at the past $\MarkovLag$ time stamps $t_{j-1}, \ldots, t_{j-\MarkovLag}$. The probability $P(Y_j = y_j|Y_{j-1} = y_{j-1},\ldots,Y_{j-\MarkovLag} = y_{j-\MarkovLag})$ is also known as the transition probability, denoted by $P(y_j| y_{j-1}, \ldots, y_{j-\MarkovLag})$ in the following context. 

For rainfall processes, the countable state space of $Y_j$ consists of $2^{\nSites}$ states, corresponding to all the combinations of the wet (rain) and dry (no rain) scenarios at $\nSites$ stations. To illustrate the construction of a Markov state sequence, consider the rainfall data at $\nSites=2$ locations recorded in  Table \ref{table:clustering_example_rainfall}. Also shown is the Markov state to which the observations at each time stamp are mapped. There are a total of 4 states, i.e., $Y_j \in \{0, 1, 2, 3\}$. We set $Y_j=0$ if no rain is observed at both locations, $Y_j=1$ if only the first location experiences rainfall, $Y_j=2$ if only the second location experiences rainfall, and $Y_j=3$ if both locations receive rain.

\begin{table}[h]
\centering
\begin{tabular}{c|c|c|c}
Time stamp & Markov state      & $x_1(t)$    & $x_2(t)$     \\ \hline
$t_1$      & 0        & 0        & 0      \\
$t_2$      & 1        & 10.54    & 0     \\
$t_3$      & 2        & 0      & 1.32       \\
$t_4$      & 3        & 2.28      & 1.63       \\
$t_5$      & 0       & 0     & 0      \\
$\vdots$   & $\vdots$ & $\vdots$ & $\vdots$ \\
$t_{\nTimes}$      & 1        & 12.21    & 0    
\end{tabular}
\caption{Illustrated mapping between the observed rainfall data and the corresponding Markov states for $\nSites=2$ locations. Observations are mapped to the Markov state depending on which location experiences rainfall.}
\label{table:clustering_example_rainfall}
\end{table}

The Markov state sequence $Y_j, j = 1, \ldots, n,$ can be fully characterized by the order $\MarkovLag$ and the transition probability $P(y_j| y_{j-1}, \ldots, y_{j-\MarkovLag})$, $j \ge \MarkovLag + 1$. Likelihood measures such as the Akaike information criterion (AIC) \cite{markov_aic} or the Bayesian information criteria (BIC) \cite{markov_bic} can be employed to estimate $\MarkovLag$. The corresponding transition probability, represented in matrix form, can then be estimated given the realizations of $Y_1, \ldots, Y_{\nTimes}$ \cite{transition_prob_est}. Based on \eqref{implication of Markov}, a sample sequence of $Y_j$ is simulated in a sequential manner for time stamps $t_{\MarkovLag+1},\dots,t_{\nTimes}$. Given a realization $y_1,\ldots,y_p$ of $Y_1,\ldots,Y_p$, we initialize $\ty_1, \ldots, \ty_p$ by using $y_1, \ldots, y_p$. Subsequently, we simulate a realization $\ty_{\MarkovLag+1}$ from the transition probability $P(y|y_1,\ldots,y_p)$. This is then used to simulate a realization $\ty_{\MarkovLag+2}$ from the transition probability $P(y|y_2,\ldots,y_{\MarkovLag},\ty_{\MarkovLag+1})$. This procedure is repeated to produce $\ty_{\MarkovLag+1},\dots,\ty_{\nTimes}$.

\subsubsection{Resampling $\nSites$-variate stochastic processes based on univariate Markov sequences}
\label{subsubsec:prelim:resampling_univariate_Markov}

In the next step, preliminary synthetic realizations of the rainfall sequence $\tbfx(t_1), \ldots, \tbfx(t_{\nTimes})$ are generated by resampling subsequences of rainy sequences present in the series $\ty_1, \ldots, \ty_{\nTimes}$. A rainy sequence is defined as the sequence of Markov states such that at least one station experiences rain for consecutive time stamps. For the case where $\nSites =2$, the rainy sequence takes values from set $\{1, 2,3\}$ which excludes 0 since it is a dry scenario. In Table~\ref{table:clustering_example_rainfall}, the subsequence $1, 2, 3$ at time stamps $t = t_2,t_3,t_4$  is a rainy sequence.

The resampling procedure is based on the bootstrap algorithm \cite{bootstrap}. Suppose we have a synthetic realization $\ty_{1},\dots,\ty_{\nTimes}$ from Section~\ref{subsubsec:prelim:markov_sequence}. For each rainy sequence present in this series, we seek an identical rainy sequence among the Markov state sequences of the observed data. The corresponding rainfall measurements of the identical rainy sequence are then  used as the synthetic rainfall values. 
For example, suppose we have a synthetic Markov state sequence of $1, 2, 3$ at the consecutive time stamps $t_{j}, t_{j+1}, t_{j+2}, j \in \{1,\dots,\nTimes-2\}$. Since it matches the existing rainy sequence shown in Table~\ref{table:clustering_example_rainfall}, we have the synthetic realization $\tbfx(t_{j}) = [\tilde{x}_1(t_{j}), \tilde{x}_2(t_{j})]^T = [10.54, 0]^T$, $\tbfx(t_{j+1}) = [0, 1.32]^T$, and $\tbfx(t_{j+2}) = [2.28, 1.63]^T$.  When there are multiple matches, we randomly choose one in a uniform manner.  The resampling procedure is repeated for all the synthetic rainy sequences present in $\ty_1, \ldots, \ty_{\nTimes}$. This results in the synthetic realization $\tbfx(t_1), \ldots, \tbfx(t_{\nTimes})$ of length $\nTimes$.

There are two limitations of resampling. First, we may not find a rainy sequence from the existing realizations that matches the synthetic Markov state sequence. This is especially the case when $\nSites$ is large which then implies that the Markov state space dimension is large. Special techniques such as divide-and-conquer need to be applied, where the synthetic rainy sequence is divided into subsequences to be matched. However, this results in inconsistencies in the statistical properties, e.g., auto-correlation functions, of the resulting realizations. Second, resampling is unable to generate unobserved values of $\bfX(t)$ as it is performed via bootstrapping. This becomes a problem if interest is on extreme events, for example. The latter limitation is addressed by the reshuffling technique described in the following subsection. However, the former still remains a limitation that hinders the scalability of this stochastic generator for large $\nSites$.

    %\TODO{Wayne: as we discussed, the reason why we start with 1 and not $\MarkovLag+1$ is due to stationarity. Should this be mentioned explicitly or is this obvious in the community? No. this is not because of stationarity. This is because the rainy sequence may start before \MarkovLag and end after \MarkovLag. But we have stated clearly that the resampling is done starting $t_1$. So people can refer to the original paper if they want further clarification}

%Note that $j$ is not necessary to be 2, which is the same as the observations.

\subsubsection{Reshuffling realizations of $\nSites$-variate stochastic processes}
\label{subsubsec:prelim:reshuffling}

To ensure the inclusion of the unobserved values of $\bfX(t)$, especially the unprecedented extremes, we perform a reshuffling procedure for each station. First, new samples are simulated from the marginal distribution for each station. These are then reshuffled according to the ranks of the realizations resulting from the resampling step in Section~\ref{subsubsec:prelim:resampling_univariate_Markov}.

In \cite{precipitation_paper}, the authors suggest to characterize the marginal distributions of $\bfX(t)$ with parametric forms, e.g., Weibull distribution for the positive parts of the distributions, calibrated to the existing rainfall observations. For each spatial location $i=1, \dots, \nSites$, denote by $\tbfx_i = [\tilde{x}_i(t_1), \ldots, \tilde{x}_i(t_{\nTimes})]$ the sequence of a resulting realization from the resampling procedure described in Section~\ref{subsubsec:prelim:resampling_univariate_Markov} and let $\tilde{\bfr}_i = [\tilde{r}_i(t_1), \ldots, \tilde{r}_i(t_{\nTimes})]$ be the corresponding ranks of the components of $\tbfx_i$, sorted in descending order. Let $t_{j_1}, \dots, t_{j_{\nTimes}}$ be the time indices, $j_1, \dots, j_{\nTimes} \in \{1, \dots, \nTimes\}$, such that $\tilde{x}_i(t_{j_1}) \ge \dots \ge \tilde{x}_i(t_{j_{\nTimes}})$. Thus, $\tilde{r}_i(t_{j_1}) = 1$ for the time stamp $t_{j_1}$ while $\tilde{r}_i(t_{j_{\nTimes}}) = \nTimes$ for the time stamp $t_{j_{\nTimes}}$. Denote by $\bfz_i = [z_i^1, \ldots, z_i^{\nTimes}]$, $z_i^1 \geq z_i^2 \geq \cdots \geq z_i^{\nTimes},$ a sorted sequence of $\nTimes$ new samples simulated from the marginal distribution of $X_i(t)$. The reshuffled sequence $\fsx_i = [\hatx_i(t_1), \ldots, \hatx_i(t_{\nTimes})]$ is then defined by 
\begin{equation}
\label{eq:reshuffle}
    \hatx_i(t_j) = z_i^{\tilde{r}_i(t_j)}, \,\, j=1,\dots,\nTimes.
\end{equation}

To illustrate, consider a 2-station rainfall process from the resampling step with a time horizon of 5 steps as shown in Table \ref{table:reshuffing_example}, where the rainfall sequences at the two stations are denoted by $\tbfx_1$ and $\tbfx_2$. The time sequence at the first location $\tbfx_1 = [2.14, 6.36, 0.64, 4.05, 1.31]$ results in a rank sequence $\tilde{\bfr}_1=[3, 1, 5, 2, 4]$ while the time sequence at the second location $\tbfx_2 = [0.51,3.24,2.46,0.60,2.00]$ results in $\tilde{\bfr}_2 = [5,1,2,4,3]$. If the sorted simulated sequences for the locations are $\bfz_1 = [4.68, 4.34, 2.58, 1.76, 1.26]$ and $\bfz_2 = [5.53, 5.27, 4.34, 2.75, 1.52]$, respectively, the reshuffled sequences are then $\fsx_1=[2.58, 4.68, 1.26, 4.34, 1.76]$ and $\fsx_2=[1.52, 5.53, 5.27, 2.75, 4.34]$ following \eqref{eq:reshuffle}. 
    
\begin{table}[!htb]
    \begin{subtable}{.4\linewidth}
      \centering
        \caption{Samples from resampling}
        \begin{tabular}{c|c|c|c|c}
            Time stamp & $\tbfx_1$ & $\tbfx_2$ & $\tilde{\bfr}_1$ & $\tilde{\bfr}_2$ \\ \hline
            $t_1$      & 2.14     & 0.51     & 3        & 5        \\
            $t_2$      & 6.36     & 3.24     & 1        & 1        \\
            $t_3$      & 0.64     & 2.46     & 5        & 2        \\
            $t_4$      & 4.05     & 0.60     & 2        & 4        \\
            $t_5$      & 1.31     & 2.00     & 4        & 3       
        \end{tabular}
    \end{subtable}%
    \begin{subtable}{.25\linewidth}
      \centering
        \caption{Sorted simulated samples}
        \begin{tabular}{c|c}
            $\bfz_1$ & $\bfz_2$ \\ \hline
            4.68     & 5.53     \\
            4.34     & 5.27     \\
            2.58     & 4.34     \\
            1.76     & 2.75     \\
            1.26     & 1.52    
            \end{tabular}
    \end{subtable}%
    \begin{subtable}{.4\linewidth}
      \centering
        \caption{Samples after reshuffling}
        \begin{tabular}{c|c|c|c|c}
            Time stamp & $\fsx_1$ & $\fsx_2$ & $\tilde{\bfr}_1$ & $\tilde{\bfr}_2$ \\ \hline
            $t_1$      & 2.58     & 1.52     & 3        & 5        \\
            $t_2$      & 4.68     & 5.53     & 1        & 1        \\
            $t_3$      & 1.26     & 5.27     & 5        & 2        \\
            $t_4$      & 4.34     & 2.75     & 2        & 4        \\
            $t_5$      & 1.76     & 4.34     & 4        & 3       
        \end{tabular}
    \end{subtable}%
\caption{Reshuffling of a hypothetical 2-station example with 5 time stamps. (a) Samples of $\tbfx(t)$ at $t_1, \ldots, t_5$ are generated from the resampling step described in Section~\ref{subsubsec:prelim:resampling_univariate_Markov}, with $\tilde{\bfr}_1$ and $\tilde{\bfr}_2$ being the corresponding ranks; (b) Synthetic samples $\bfz_1$ and $\bfz_2$ are simulated from the marginal distribution at each location; (c) Samples are reshuffled according to the ranks $\tilde{\bfr}_1$ and $\tilde{\bfr}_2$.}
\label{table:reshuffing_example}
\end{table}    

Although we introduce and illustrate the reshuffling above on a single realization of rainfall process of length $\nTimes$, it is suggested that the reshuffling procedure at each location  be conducted over a relatively long time duration \cite{precipitation_paper}, e.g., over concatenated multiple synthetic realizations. This is because longer time series after reshuffling better resemble the original time series. The reshuffling approach matches the marginal distributions exactly and provides a satisfactory approximation for other statistical properties \cite{precipitation_paper}. 

%     the reshuffling approach is adopted, which is conducted per station and consists of two steps. The new samples are firstly originated from the marginal distributions, then get reshuffled per station according to the ranks of the realizations resulted from the resampling step.

% while the resulting time series is approximately stationary and .

% This procedure is carried out similarly to obtain $\tilde{\textbf{x}}_2$. 

% , which indicates that $\tilde{x}_1(t_1)$ corresponds to the third largest element of the new samples $y^1_3$, $\tilde{x}_1(t_2)$ is equal to the largest element $y^1_1$, etc

% \begin{figure}[h]
% \centering
% \includegraphics[width=15cm]{figures/Reshuffling_example.png}
% \caption{Reshuffling of a hypothetical 2-location example with 5 time stamps: As a start, we have the values of $\textbf{x}_1$ and $\textbf{x}_2$ at $t_1, \ldots, t_5$ from previous steps, where $\textbf{r}_1$ and $\textbf{r}_2$ are the corresponding ranks (a); Next, synthetic values $\textbf{y}^1$ and $\textbf{y}^2$ are then sampled from the distribution at each location (b); Finally, the samples are reshuffled according to the ranks  $\textbf{r}_1$ and $\textbf{r}_2$ (c).}
% \label{fig:reshuffing_example}
% \end{figure}

\subsection{Deep learning models for time series forecasting}
\label{subsec:prelim:deep_learning_model}
% \subsection{Model architecture}

\begin{figure}[h]
\centering
\includegraphics[width=14cm]{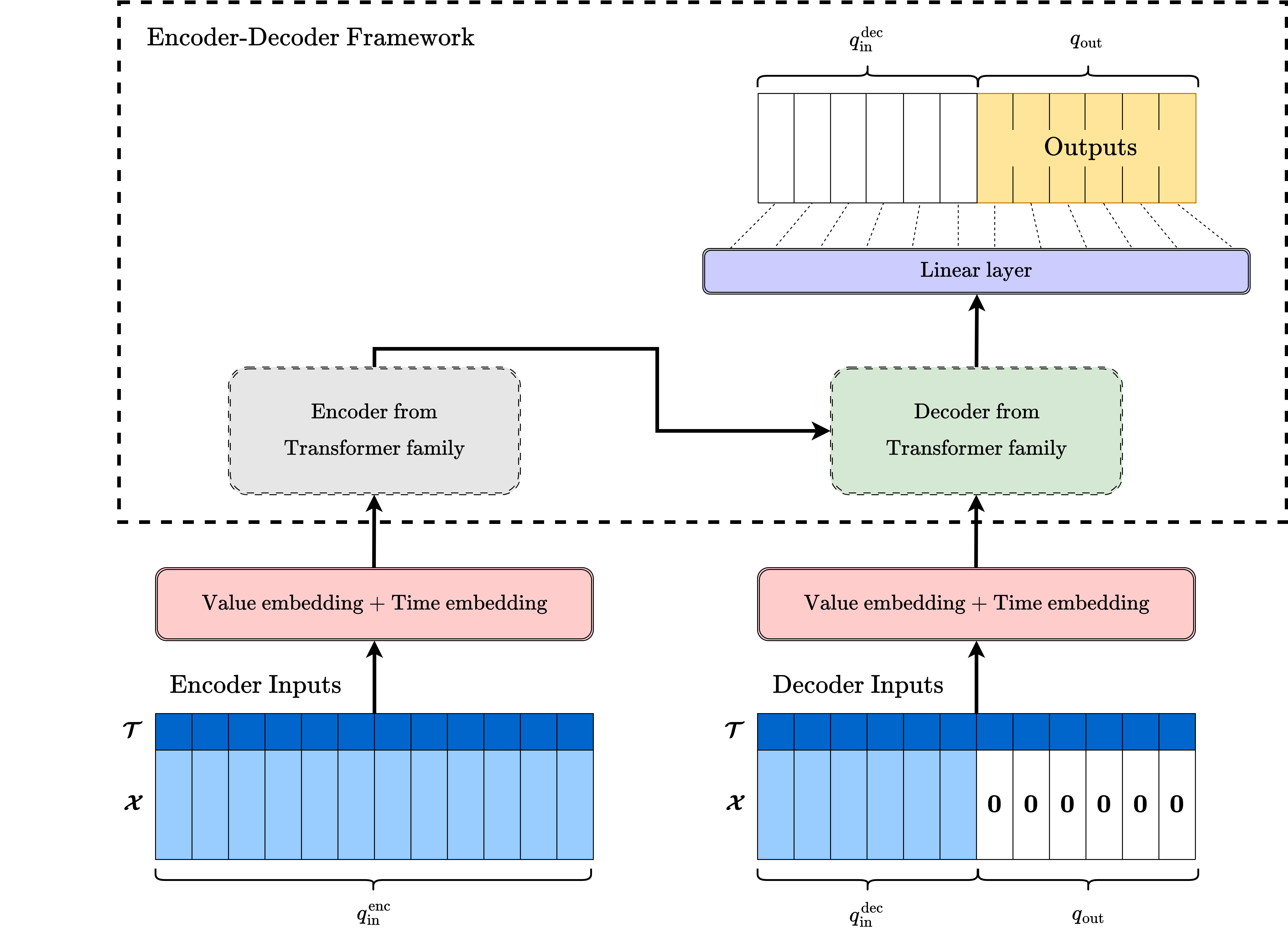}
\caption{Deep learning model architecture based on the encoder-decoder framework. The model processes inputs through an embedding layer (red block), generating the hidden representation which undergoes further updates in the encoder layers (gray block). The decoder (green block), in conjunction with a linear layer (purple block), utilizes the hidden representation from the encoder for generative inference, yielding the predicted sequence (highlighted in yellow).}
\label{fig:model_architecture} 
\end{figure}

Deep learning models for time series forecasting mirror models developed in the natural language processing (NLP) domain due to the sequential nature of both tasks. Recently, commonly-adopted models in NLP typically follow the Transformer architecture which has an encoder-decoder structure \cite{encoder_decoder_1, encoder_decoder_2, encoder_decoder_3}. The model architecture is shown in Figure~\ref{fig:model_architecture}. Within this structure, the encoder captures the dependencies and patterns inherent in the input sequence, subsequently conveying the extracted information to the decoder, which is tasked with generating predictions.

Let $\bfTcal^{\text{enc}} = [t_1,\dots,t_{\qInEnc}]$ be a vector of increasing time stamps of length $\qInEnc < \nTimes$ and $\bfXcal^{\text{enc}} \in \mathbb{R}^{\nSites \times \qInEnc}$ be the time series matrix where each row represents the stochastic process $X_i(t)$ for the $i^{\text{th}}$ location, $i = 1, \ldots, \nSites$, at the time stamps indicated in $\bfTcal^{\text{enc}}$. Given $\bfXcal^{\text{enc}}$ and $\bfTcal^{\text{enc}}$, the deep learning model predicts the subsequent sequence $\bfXcal^{\text{out}} \in \mathbb{R}^{\nSites \times \qOut}$ at $\qOut$ time stamps specified in $\bfTcal^{\text{out}}$. The model proceeds by passing the inputs $\bfXcal^{\text{enc}}$, $\bfTcal^{\text{enc}}$ through the embedding layer (red block) to obtain the hidden representation $\bfZcal^{\text{enc}}$. The hidden representation refers to the output matrices of the hidden layers, which all have dimension $\dmodel$, a hyperparameter determining the length of the hidden elements. The hidden representation is then updated through the layers of the encoder (gray block). Based on the resulting representation matrix $\bfZcal^{\text{enc}}$ and $\bfTcal^{\text{out}}$, the decoder (green block) combined with a linear layer (purple block) performs generative inference to produce the output sequence $\bfXcal^{\text{out}}$ (highlighted in yellow). 

In the following two subsections, we detail the embedding layer and the encoder-decoder structure, which are the two major components of this Transformer-based model.

% The encoder encapsulates the dependencies and patterns of the input sequence which then passes this information to the decoder which is responsible for generating predictions.

\subsubsection{Embedding layer}
The embedding layer serves to convert the inputs into a hidden representation which is then utilized by the encoder and decoder. It consists of two components, namely, the value embedding and time embedding, corresponding to the time series input $\bfXcal \in \mathbb{R}^{\nSites \times q}$ and the time sequence $\bfTcal$ of length $q$, respectively.  

The value embedding is a 1D-convolutional layer \cite{conv_layer} with specified kernel size and circular padding to map the spatial dimension $\nSites$ to the hidden dimension $\dmodel$. We denote the operation of the value embedding applied to an input $\bfXcal$ by ValueEmbedding($\bfXcal$) which produces a hidden representation of dimension $\dmodel \times q$.  

The time embedding, on the other hand, depends on whether or not the time argument is unitless. If the time argument is unitless, only the order of the sequence matters. Therefore, the positional embedding introduced in \cite{attention_paper} is adopted as the time embedding. Given a time sequence of length $q$, the positional embedding returns a matrix of dimension $\dmodel \times q$, where the $(2k, j)$ and $(2k+1, j)$ entries, $k = 0, \ldots, \lfloor \dmodel / 2 \rfloor-1$, $j = 0, \ldots, q-1$, are defined as $\sin(j/(10000 ^{2k / \dmodel}))$ and $\cos(j/(10000 ^{2k / \dmodel}))$, respectively. In contrast, if the time argument contains unit information, i.e., the data is in the format of year-month-day-hour-minute-second or a subset of any of these units, \cite{informer_paper} uses the time feature embedding instead to take the temporal information into consideration. To illustrate, if the time sequence is of length $q$ and data is recorded in the year-month-day-hour format, we reshape the $q$-dimensional time sequence into a $4 \times q$ matrix where each row records the standardized values for each unit of measure. A linear layer without bias is then applied to map the $4 \times q$ matrix to a $\dmodel \times q$ matrix.  We denote the operation of the time embedding applied to an input $\bfTcal$ by TimeEmbedding($\bfTcal$) which produces a matrix of dimension $\dmodel \times q$.

The output representation $\bfZcal$ from the embedding layer is thus defined as 
\begin{equation}
\bfZcal = \text{ValueEmbedding}(\bfXcal)  + \text{TimeEmbedding}(\bfTcal). \label{eq:embedding}
\end{equation}

\subsubsection{Encoder-decoder framework}

The encoder-decoder framework is commonly used for sequence-to-sequence tasks, e.g., time series forecasting and machine translation in the NLP domain. The work \cite{attention_paper} has shown the effectiveness of adopting the self-attention mechanism as the primary building block in the encoder-decoder framework. Self-attention processes a sequence by replacing each element by a weighted average of the rest of the sequence so that the dependencies of each element with respect to the others in the sequence are learned. The encoder encapsulates all the information about the input sequence through multiple layers of self-attention and its output is fed to the decoder for generating predictions.

Self-attention, first proposed in \cite{attention_paper}, maps a query and a set of key-value pairs to obtain an output representation where the query, key, and value matrices stem from the input representation $\bfZcal \in \mathbb{R}^{\dmodel \times q}$. Mathematically, self-attention is defined as 
\begin{align}
\text{Self-Attention}(\bfZcal) &= \bfV~\text{Softmax}\left(\frac{\bfQ^T \bfK}{\sqrt{\dmodel}} \right), \label{eq: attention} \\ 
\bfQ &= \bfW_Q \bfZcal, \nonumber \\
\bfK &= \bfW_K \bfZcal,  \nonumber \\
\bfV &= \bfW_V \bfZcal,  \nonumber
\end{align}
where $\bfW_Q, \bfW_K, \bfW_V \in \mathbb{R}^{\dmodel \times \dmodel}$ are the query, key, and value conversion matrices,  $\text{Softmax}(\cdot)$ denotes the softmax function defined in \cite{Goodfellow-et-al-2016}, and $\bfQ, \bfK, \bfV \in \mathbb{R}^{\dmodel \times q}$ are the resulting query, key, and value matrices. As seen in \eqref{eq: attention}, the output representation of the self-attention layer can be regarded as the weighted sum of the input values. For simplicity, we set the first dimension of the query and key matrices to be $\dmodel$, but in general, they only need to be compatible with each other. % (not necessary to be $\dmodel$). 

% that assigns to each value the normalized weight computed by the scaled dot product of the query with the corresponding key

The attention mechanism introduced above is single-headed as it only performs the mapping from the $\dmodel$-dimensional query to the $\dmodel$-dimensional key-value pairs once. As suggested in \cite{attention_paper}, it is beneficial to adopt a multi-head mechanism, i.e., we perform $\nHeads$ such mappings in parallel with $\dmodel / \nHeads$-dimensional query and key-value pairs. The resulting output representations from all independent mappings are concatenated and further linearly projected to match the output dimension $\dmodel$. More advanced attention mechanisms have been developed in the deep learning community to improve its performance. For example, the probsparse-attention is developed for Informer \cite{informer_paper} with the aim of significantly increasing the efficiency of the attention mechanism. The attention is distilled such that each key can only attend to the dominant queries. The Autoformer \cite{autoformer_paper} utilizes auto-correlation-attention instead of the standard attention mechanism. It introduces the notion of sub-series similarity based on the series periodicity and aggregates similar sub-series from underlying periods.

% overcome the limitations of the aforementioned structure and  (I'm deleting this now because I realized that we didnt really talk about the limitations)

The encoder is designed to learn and extract the dependencies and patterns of the input sequence $\bfXcal =\bfXcal^{\text{enc}}$ across the temporal and spatial dimensions. It is composed of a stack of $\nEnc$ identical blocks, generally consisting of two sub-layers each. The first is a multi-head attention layer described above while the second is a fully-connected feed-forward network with relu activation function \cite{relu_gelu} and can be mathematically expressed as 
\begin{equation}
\text{FFN}(\bfZcal) = \bfW_2 \max(0, \bfW_1 \bfZcal + \bfb_1) + \bfb_2, \label{eq:feed-forward}
\end{equation}
where $\bfZcal$ is the input hidden representation with $q = \qInEnc$, $\bfW_1 \in \R^{\dff \times \dmodel}$ and $\bfW_2 \in \R^{\dmodel \times \dff}$ are weight matrices with
$\dff$ being the dimension of the feed-forward layer, and $\bfb_1 \in \R^{\dff \times q}$ and $\bfb_2 \in \R^{\dmodel \times q}$ are the bias matrices. Each sub-layer is succeeded by a layer normalization \cite{layer_normalization}. Special techniques such as distillment, decomposition, etc, may be applied in between sub-layers depending on the choice of the attention mechanism. The final hidden representation of the encoder is then fed to the decoder.
% with identical columns
    
On the other hand, the aim of the decoder is to perform generative inference on the output sequence $\bfXcal^{\text{out}}$ based on the time sequences $\bfTcal^{\text{out}}$. Similar to NLP applications wherein we apply a start token for dynamic decoding \cite{bert}, we initiate the inference with  $\bfXcal^{\text{dec}}_{\text{start}} \in \R^{\nSites \times \qInDec}$ and $\bfTcal^{\text{dec}}_{\text{start}}\in \R^{\qInDec}$ which are subsets of $\bfXcal^{\text{enc}}$ and $\bfTcal^{\text{enc}}$, respectively, representing the last $\qInDec$ columns of the aforementioned matrices. The input of the decoder is the embedding \eqref{eq:embedding} applied to the following matrices: 
\begin{align}
     \bfXcal = \bfXcal^{\text{dec}} &= \text{Concat}(\bfXcal^{\text{dec}}_{\text{start}}, \textbf{0}), \\
     \bfTcal = \bfTcal^{\text{dec}} &= \text{Concat}(\bfTcal^{\text{dec}}_{\text{start}}, \bfTcal^{\text{out}}) \notag, \label{eq: decoder_input}
\end{align}
where $\text{Concat}(\cdot)$ denotes the matrix concatenation operation, $\textbf{0} \in \R^{\nSites \times \qOut}$ is a zero matrix which serves as the placeholder of the output sequence $\bfXcal^{\text{out}}$, $\bfXcal^{\text{dec}} \in \R^{\nSites \times (\qInDec + \qOut)}$, and $\bfTcal^{\text{dec}} \in \R^{(\qInDec + \qOut)}$. The structure of the decoder is similar to that of the encoder, which is composed of a stack of $\nDec$ identical blocks. In addition to the two sub-layers in each block, namely the attention layer and the feed-forward layer, the decoder includes a third sub-layer which performs multi-head cross-attention on the output of the encoder stacks to incorporate the learned dependencies and patterns of the input sequence. The cross-attention mechanism is a variant of the self-attention mechanism described above. It has the same structure as self-attention, except that the input representation to the key and value matrices is the output from the encoder instead of the output from the previous block in the decoder. Therefore, we have $\bfQ = \bfW_Q \bfZcal^{\text{dec}} \in \mathbb{R}^{\dmodel \times (\qInDec + \qOut)}$, $\bfK = \bfW_K \bfZcal^{\text{enc}} \in \mathbb{R}^{\dmodel \times \qInEnc}$, and $\bfV = \bfW_V \bfZcal^{\text{enc}} \in \mathbb{R}^{\dmodel \times \qInEnc}$. The output representation from the cross-attention layer and that of the decoder has dimension $\dmodel \times (\qInDec + \qOut)$.  It is then transformed to have size $\nSites \times (\qInDec + \qOut)$ via a linear layer with bias. Only the last $\qOut$ positions of the output sequence are of interest. Note that the inference procedure under the decoder is conducted in a single forward step instead of in an auto-regressive manner.

\subsection{Problem formulation}

Given the realizations $\bfx(t)$ of the stationary and ergodic process $\bfX(t), t \in [0, \timeEndObs],$ with continuous marginal distributions, we aim to construct a stochastic generator to generate additional samples $\fsx(t)$ of $\bfX(t)$ for $t \in [0, \timeEndSim]$ that preserve statistical properties of the existing realizations $\bfx(t)$. These synthetic realizations can then be employed in various downstream applications.

In this work, we consider the case where the marginal distributions of the components of $\bfX(t)$ are Gaussian without loss of generality. If the observed data $\bfx(t)$ of $\bfX(t)$ have components with non-Gaussian marginals, a transformation can be applied to the data during pre-processing. For each location $i$ with marginal cumulative distribution function (CDF) $F_i$, the transformed observations $x_i(t)$ are given by the invertible mapping $\Phi^{-1}[F_i(x_i(t))]$, where $\Phi$ is the standard Gaussian CDF. The marginal distribution $F_i$ can be characterized by parametric distributions as in \cite{precipitation_paper}, by its empirical distribution, or by a mixture of both \cite{translation_paper_mix}.

We propose a stochastic generator that combines and extends the two models introduced in Sections~\ref{subsec:prelim:stochastic_generator} and~\ref{subsec:prelim:deep_learning_model}. In particular, the Transformer-based deep learning model for time series forecasting is adopted as a critical component of this generator, facilitating scalability in scenarios where the number of spatial locations is large and the simulation horizon is long.

\section{Stochastic generator for $\nSites$-variate stochastic processes using deep learning models}
\label{sec:newmodel}

We present GenFormer, a stochastic generator for $\nSites$-variate stochastic processes using deep learning models. It extends the stochastic generator described in Section~\ref{subsec:prelim:stochastic_generator} in the following three aspects. First, we provide a generalized approach to define the state space of the univariate Markov sequence $Y_j$ by using $K$-means clustering, thereby extending the applicability of the model in Section~\ref{subsubsec:prelim:markov_sequence} beyond precipitation data. Second, we replace the resampling procedure in Section~\ref{subsubsec:prelim:resampling_univariate_Markov} by a deep learning model which constitutes the mapping from the Markov states to the inferred values of the \nSites-variate process. The deep learning model has the same encoder-decoder framework used for time series forecasting in Section~\ref{subsec:prelim:deep_learning_model} but with an additional embedding for the Markov states in the embedding layer. The deep learning model serves to improve the scalability of the resampling procedure. However, the fidelity of the resulting mapping in approximating statistical properties of interest depends on the performance of the deep learning model. Thus, to preserve statistical properties such as the spatial correlation and marginal distributions, we include a model post-processing step based on Cholesky decomposition and the reshuffling technique in Section~\ref{subsubsec:prelim:reshuffling} as the third extension. It is also worth-noting that since simulating univariate Markov sequences can become challenging for high Markov orders, an additional light-weight deep learning model is proposed to address this limitation.

The resulting synthetic data produced by GenFormer is able to exactly match the target marginal distributions and approximately match the second-moment properties. Moreover, it can also capture the higher-order statistical properties of quantities of interest derived from $\bfX(t)$. Most importantly, GenFormer can be applied to stochastic processes in which the number of locations is large and the simulation horizon is long.

The proposed approach is comprised of two stages, namely model construction and model simulation, described in Sections~\ref{subsec:newmodel:model_construction} and~\ref{subsec:newmodel:model_simulation}, respectively. In Section~\ref{subsec:newmodel:model_construction}, we focus on model training and computational aspects of the univariate Markov sequence generator based on $K$-means clustering and the deep learning model with Markov state embedding. In Section~\ref{subsec:newmodel:model_simulation}, we discuss how synthetic realizations can be generated using the aforementioned models coupled with the post-processing procedure. Section~\ref{subsec:newmodel:simulation_algorithm_summary} summarizes the proposed GenFormer algorithm. 

    % except that an additional embedding for the Markov states is included in the embedding layer to incorporate the corresponding information (which is not present in the context of time series forecasting). 

\subsection{Model construction}
\label{subsec:newmodel:model_construction}

\subsubsection{Construction of univariate Markov sequence via clustering}
\label{subsubsec:newmodel:markov_sequence_clustering}

Suppose we have $\nTimesObs$ time stamps $t_1, \ldots, t_{\nTimesObs}$ that evenly partition the duration $[0, \timeEndObs]$ of the observed data. The countable state space of the univariate Markov sequence $Y_1, \ldots, Y_{\nTimesObs}$ of the stochastic generator proposed in \cite{precipitation_paper}  contains all the combinations of the wet and dry scenarios of the $\nSites$ stations, resulting in $2^{\nSites}$ Markov states. In general, there is no established rule to define the state space. We thus propose to partition the realizations $\bfx(t)$ of $\bfX(t)$ into subsets of interest, each indexed by a positive integer, which represents certain spatial variation of $\bfX(t)$.

We achieve this through $K$-means clustering \cite{kmeans}  which is a commonly-used unsupervised learning algorithm to efficiently partition a set of vectors into distinct and non-overlapping clusters based on inherent similarities. It can thus be employed to segregate the set $\{\bfx(t_1),\dots,\bfx(t_{\nTimesObs})\}$ of realizations at $\nTimesObs$ time stamps into $\nClusters$ clusters, where $\nClusters$ is a prescribed hyperparameter. Each cluster is represented by a centroid. The clustering algorithm is an iterative process of sample reassignment and centroid recalculation with the goal of minimizing within-cluster variance while maximizing between-cluster distance. The algorithm is sensitive to the initial choice of centroids and has to be performed with multiple sets of starting points. It results in each observation being allocated to a centroid which then corresponds to a state of the univariate Markov sequence.

We illustrate the application of $K$-means clustering in Table~\ref{table:clustering_example}. Consider a 3-variate stochastic process $\bfX(t)$ with 5 time stamps. The components $X_1(t)$, $X_2(t)$, and $X_3(t)$ are highly-correlated with each other and all follow a bimodal distribution centered at $12$ and $2$. We set $\nClusters = 2$. By applying $K$-means clustering, the Markov state $y_j$ is assigned to 1 and 2 at time stamps $t_j$ when the realizations $\bfx(t_j)$ are near the modes $12$ and $2$, respectively. 

\begin{table}[h]
\centering
\begin{tabular}{c|c|c|c|c}
Time stamp & $y$       & $x_1(t)$    & $x_2(t)$    & $x_3(t)$    \\ \hline
$t_1$      & 1        & 11.32    & 10.12    & 12.56    \\
$t_2$      & 1        & 10.54    & 11.98    & 14.12    \\
$t_3$      & 2        & 0.5      & 1.32     & 2.63     \\
$t_4$      & 2        & 1.8      & 3.24     & 2.12     \\
$t_5$      & 1        & 12.21    & 12.34    & 11.79   
\end{tabular}
\caption{Illustrated mapping of Markov states to a 3-variate process based on $K$-means clustering. We have $y_j = 1$ when $x_1(t_j)$, $x_2(t_j)$, $x_3(t_j)$ are near $12$, and $y_j = 2$ when $x_1(t_j)$, $x_2(t_j)$, $x_3(t_j)$ are near $2$.}
\label{table:clustering_example}
\end{table}

Given the mapped Markov state sequence $y_1, \ldots, y_{\nTimesObs}$, the estimation of the Markov order $\MarkovLag$ along with the transition matrix is akin to the approach outlined in Section~\ref{subsubsec:prelim:markov_sequence}. Utilizing these estimates, the synthetic realizations of the Markov state sequence can be subsequently generated.

\subsubsection{Deep learning model with Markov state embedding}
\label{subsubsec:newmodel:deep_learning_model}

The resampling procedure introduced in \cite{precipitation_paper} is specifically designed for rainy sequences and suffers from the curse of dimensionality. We therefore propose to train a Transformer-based deep learning model with Markov state embedding that maps the Markov state sequence $y_1, \ldots, y_{\nTimesObs}$ to the realization of the $\nSites$-variate process $\bfx(t_1), \ldots, \bfx(t_{\nTimesObs})$. Let $\bfYcal^{\text{enc}} \in \R^{\qInEnc}$ be a vector of the Markov state sequence corresponding to the vector of increasing time stamps $\bfTcal^{\text{enc}}$ and the matrix of observations $\bfXcal^{\text{enc}}$ at the specified time stamps. Given $\bfXcal^{\text{enc}}$, $\bfTcal^{\text{enc}}$, and $\bfYcal^{\text{enc}}$, the deep learning model aims to infer the subsequent sequence $\bfXcal^{\text{out}}$ based on the corresponding Markov state sequence $\bfYcal^{\text{out}}$ specified at time stamps in $\bfTcal^{\text{out}}$. The inputs of the model to perform inference starting from time stamp $j+1$ are as follows:
    \begin{itemize}
         \item $\bfTcal^{\text{enc}} = [t_{j-\qInEnc+1}, \dots, t_j] \in \R^{\qInEnc}$, a sequence of time stamps with recorded observations;
         \item $\bfXcal^{\text{enc}} = [\bfx(t_{j-\qInEnc+1}), \dots, \bfx(t_{j})] \in \R^{\nSites \times \qInEnc}$, a matrix of observations of $\bfX(t)$ at time stamps in $\bfTcal^{\text{enc}}$;
         \item $\bfYcal^{\text{enc}} = [y_{j-\qInEnc+1}, \dots, y_j] \in \R^{\qInEnc}$, a Markov state sequence corresponding to $\bfXcal^{\text{enc}}$;
         \item $\bfTcal^{\text{out}} = [t_{j+1}, \dots, t_{j+\qOut}] \in \R^{\qOut}$, a sequence of time stamps for the inference stage;
         \item $\bfYcal^{\text{out}} = [y_{j+1}, \dots, y_{j+\qOut}] \in \R^{\qOut}$, a Markov state sequence for the inference stage.
    \end{itemize}
    The output of the model is $\bfXcal^{\text{out}} = [\bfx(t_{j+1}), \dots, \bfx(t_{j+\qOut})] \in \R^{\nSites \times \qOut}$, a time series matrix corresponding to the Markov state sequence $\bfYcal^{\text{out}}$ and time stamps in $\bfTcal^{\text{out}}$. 

\begin{figure}[h]
\centering
\includegraphics[width=14cm]{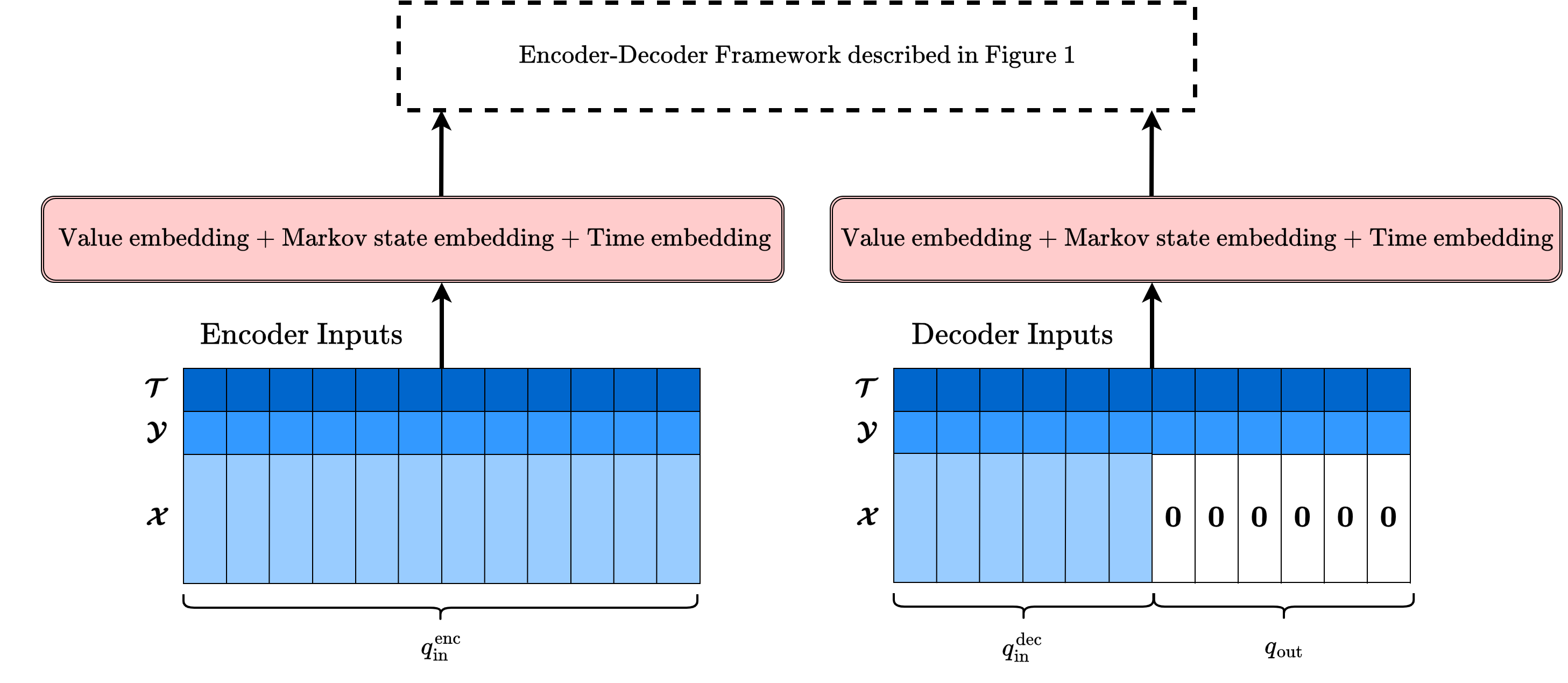}
\caption{Transformer-based deep learning model with Markov state embedding. The proposed approach includes a Markov state embedding in addition to the value and time embedding present in the embedding layer. The remainder of the model architecture is the same as in Figure~\ref{fig:model_architecture}.}
\label{fig:model_architecture_Markov_state} 
\end{figure}

In contrast to the deep learning model described in Section~\ref{subsec:prelim:deep_learning_model}, the embedding layer also needs to incorporate information from the Markov state sequence $\bfYcal$ in the hidden representation. We propose to include a Markov state embedding, wherein each Markov state is represented by a embedding vector of size $\dmodel$ learned in the training stage. The operation MarkovStateEmbedding($\bfYcal$) retrieves the respective embedding vectors in the same order as the Markov states in $\bfYcal$ via a dictionary mapping, culminating in a matrix of $\dmodel \times q$. Figure~\ref{fig:model_architecture_Markov_state} updates the architecture shown in Figure~\ref{fig:model_architecture} to include the proposed embedding for the Markov states. Given the time series input $\bfXcal$, the Markov state sequence $\bfYcal$, and the time sequence $\bfTcal$, the output representation $\bfZcal \in \R^{\dmodel \times q}$ from the embedding layer then becomes 
\begin{equation}
\bfZcal = \text{ValueEmbedding}(\bfXcal) + \text{MarkovStateEmbedding}(\bfYcal) + \text{TimeEmbedding}(\bfTcal).
\end{equation}
    
As in Section~\ref{subsec:prelim:deep_learning_model}, the model adopts the encoder-decoder framework with self-attention mechanism. The encoder is designed to extract the spatial and temporal patterns of $\bfXcal^{\text{enc}}$ and its relationship to $\bfYcal^{\text{enc}}$ and $\bfTcal^{\text{enc}}$ from the previous $\qInEnc$ time stamps. The decoder then uses the extracted representation from  the encoder to perform inference on $\bfXcal^{\text{out}}$ at subsequent time stamps with the decoder inputs
\begin{align}
     \bfXcal^{\text{dec}} &= \text{Concat}(\bfXcal^{\text{dec}}_{\text{start}}, \textbf{0}), \label{eq: decoder_input} \\
     \bfYcal^{\text{dec}} &= \text{Concat}(\bfYcal^{\text{dec}}_{\text{start}}, \bfYcal^{\text{out}}), \nonumber \\
     \bfTcal^{\text{dec}} &= \text{Concat}(\bfTcal^{\text{dec}}_{\text{start}}, \bfTcal^{\text{out}}), \nonumber
\end{align}   
where $\bfXcal^{\text{dec}}_{\text{start}}$, $\bfYcal^{\text{dec}}_{\text{start}}$, and $\bfTcal^{\text{dec}}_{\text{start}}$ are sub-matrices pertaining to the last $\qInDec$ columns of $\bfXcal^{\text{enc}}$, $\bfYcal^{\text{enc}}$, and $\bfTcal^{\text{enc}}$, respectively. The $\textbf{0}$ matrix is a placeholder for the output sequence $\bfXcal^{\text{out}}$ which is later replaced by the outputs from the decoder.

Note that the hyperparameter $\qInEnc$ is not required to be the same as the Markov order, i.e., $\qInEnc \neq \MarkovLag$ in general. Similarly, setting $\qOut = 1$ is not obligatory. In practice, we select $\qInEnc > \MarkovLag$ and $\qOut > 1$ because the deep learning model is capable of extracting temporal patterns of long time series of length $\qInEnc$, even though the Markov process is assumed to have a shorter memory of length $\MarkovLag$. Meanwhile, a larger $\qOut$ can significantly accelerate the inference by generating samples at $\qOut$ time stamps at once instead of repeating the inference $\qOut$ times. On the other hand, $\qInEnc$ and $\qOut$ cannot be too large since this may potentially hinder the accuracy of the model as will be discussed in Section~\ref{subsubsec:newmodel:model_training}. The proposed deep learning model is trained using the available realizations $\bfx(t)$ of $\bfX(t)$.

    %\item Moved from section 2.3: For simplicity, unified notations $\bfx(t)$ and $\tbfx(t)$ irrespective of their underlying marginal distributions are employed in the descriptions of overall methodology in Section.~\ref{sec:newmodel}. \TODO{Delete this!}

    % \item We propose to use the deep learning model with Markov state embedding to constitute a mapping from the Markov state sequence, i.e., $y_1, \ldots, y_{\nTimes}$, to the realization of the $\nSites$-variate process $\bfx(t_1), \ldots, \bfx(t_{\nTimes})$. This is a replacement of the resampling step, introduced in \cite{precipitation_paper}, which is specifically designed for the rainy sequences and suffers from the curse of dimensionality.    

    % filled after passing through all the blocks in the decoder. 

    % where the regularization techniques need to be applied the ensure the generalizability of the model to the new samples. We will discuss the training procedure in Section.~\ref{subsubsec:newmodel:model_training}.

\subsubsection{Deep learning model for Markov state sequence generation}
\label{subsubsec:newmodel:deep_learning_model_Markov_state_generation}

In order to infer the $\nSites$-variate process, the decoder in Section~\ref{subsubsec:newmodel:deep_learning_model} requires a synthetic realization of the Markov state sequence $\bfYcal^{\text{out}}$. Synthetic realizations of the Markov state sequence can be obtained based on the specified transition matrix with Markov order $\MarkovLag$, which can be estimated from the available realizations of the mapped Markov state sequence $y_1, y_2, \ldots, y_{\nTimesObs}$. While this is feasible for $\MarkovLag=1$, the estimation of the transition matrix when $\MarkovLag \ge 2$ may be challenging due to the exponential growth of the transition matrix dimension given by $\nClusters^{\MarkovLag} \times \nClusters$. When $\nClusters$ and $\MarkovLag$ are large, an accurate estimation of the transition matrix can be computationally intensive, or even prohibitive, and requires a significant amount of data to avoid obtaining a sparse matrix which can restrict the generation of unprecedented sequences. We address this limitation by using a light-weight deep learning model, referred to as the deep learning model for Markov state sequence generation. Such model takes the Markov states at the previous $\MarkovLag$ time stamps as the input, calculates the probability of each of the $\nClusters$ states, and samples a realization of the Markov state according to these probabilities for the next time stamp. 

\begin{figure}[h]
    \centering
    \includegraphics[width=9cm]{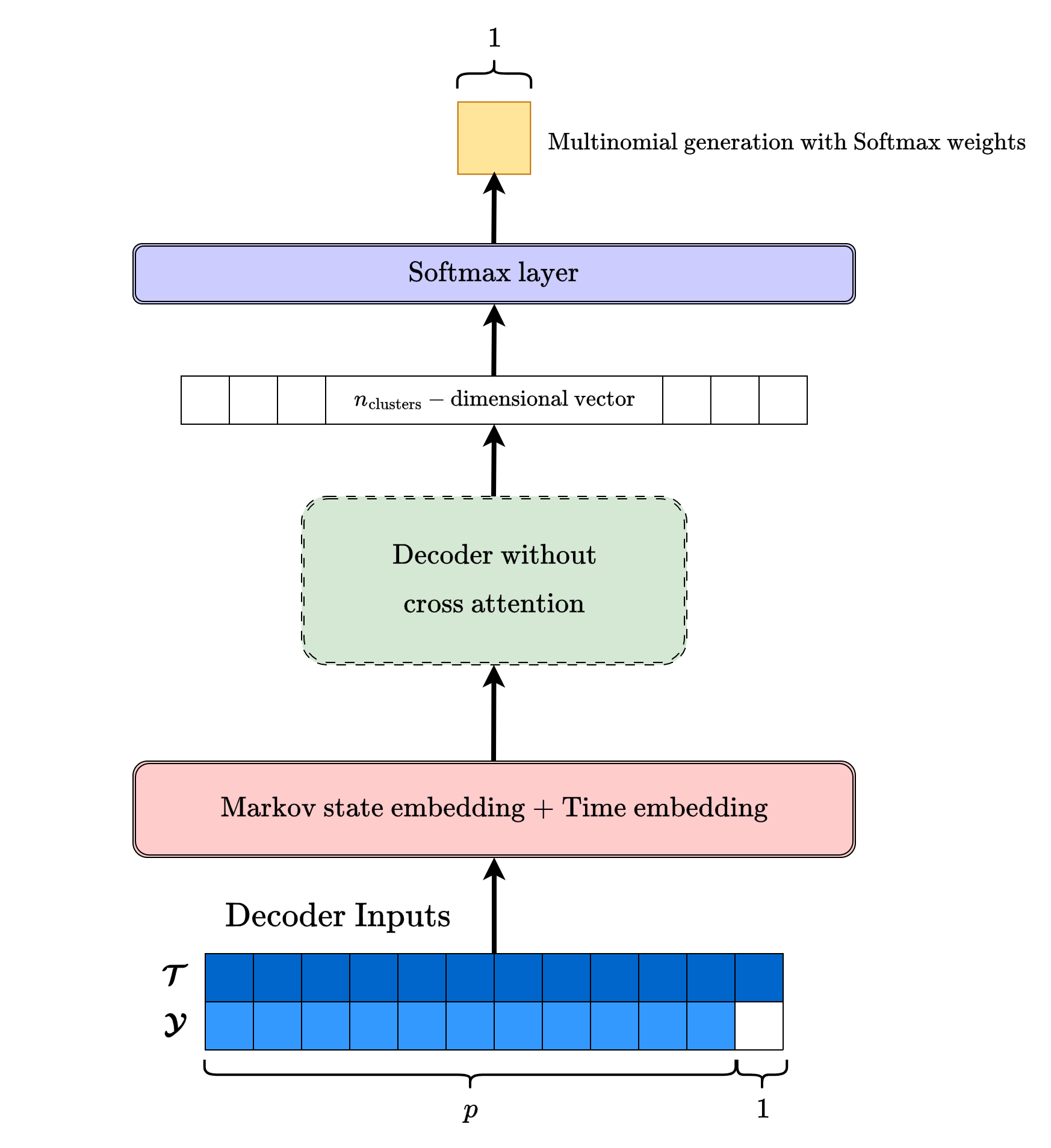}
    \caption{Deep learning model for Markov state sequence generation when Markov order $\MarkovLag \ge 2$. We adopt a decoder-only structure without cross attention mechanism. The input of the model is the Markov states in the previous $\MarkovLag$ time stamps concatenated by a vector of length 1. This is passed to an embedding layer and multiple decoder blocks. The Softmax layer normalizes the weights of Markov states to obtain probabilities which the multinomial random variable generator utilizes to generate synthetic Markov states.} 
    \label{fig:markov_model} 
\end{figure}

The architecture of the light-weight model is shown in Figure~\ref{fig:markov_model}. The model only adopts a decoder structure which takes as input the Markov states in the previous $\MarkovLag$ time stamps that is concatenated by a placeholder represented by a vector of length 1. This is followed by a Markov state embedding and time embedding (red block). The resulting hidden representation is then fed into $\nMarkov$ blocks of the decoder (green block), wherein each block contains a multi-headed attention layer and a feed-forward layer without the cross attention. This process yields a vector of size $\nClusters$ (white block), with each component representing the weight of the respective Markov state in the state space. These weights are then normalized using a softmax layer (purple block) to obtain the probability for each Markov state. Finally, a Markov state is simulated from a multinomial random variable generator based on these probabilities. The training of this model is based on the realizations of $y_1, \ldots, y_{\nTimesObs}$. 

    % \item As noted in \cite{precipitation_paper}, the Markov order $\MarkovLag$ and the corresponding transition matrix can be estimated from the realizations of the mapped Markov state sequence $y_1, y_2, \ldots, y_{\nTimes}$, which is in turn used for sampling the new synthetic sequence. This is straightforward for $\MarkovLag = 1$. However, 

     % and is also elaborated in Section~\ref{subsubsec:newmodel:model_training}. 

\subsubsection{Computational aspects of training deep learning models}
\label{subsubsec:newmodel:model_training}

Sections~\ref{subsubsec:newmodel:deep_learning_model} and \ref{subsubsec:newmodel:deep_learning_model_Markov_state_generation} discussed the architectures of the deep learning models we utilize in this work. In this subsection, we focus on the computational aspects for the practical implementation of these models.

\begin{figure}[h]
\centering
\includegraphics[width=10cm]{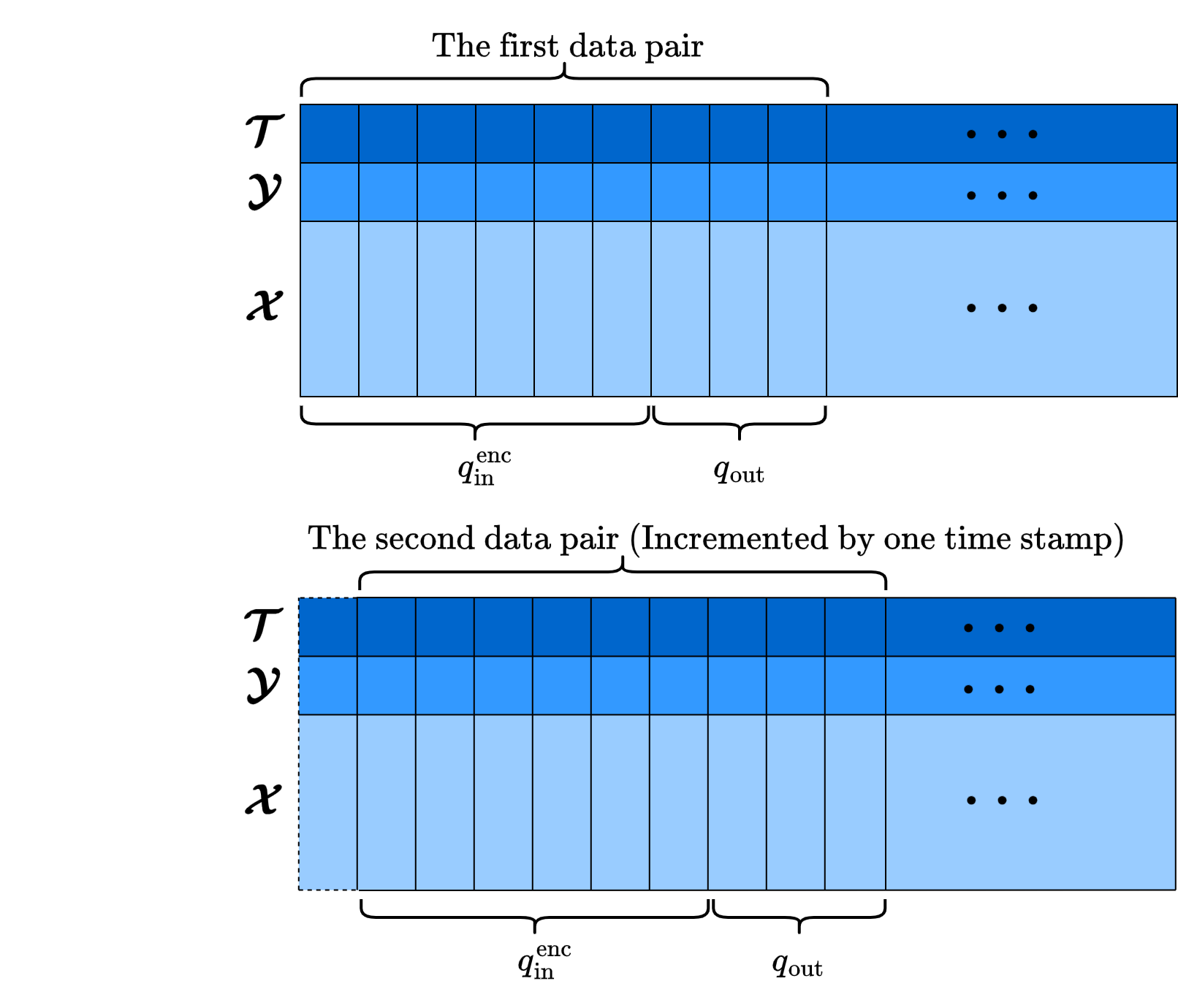}
\caption{Construction of input-output data pairs. For each sequence of realizations, we apply a sliding window of length $\qInEnc + \qOut$ to the time series matrix $\bfXcal$ and the vectors $\bfYcal$ and $\bfTcal$ of Markov state and time sequences. The first $\qInEnc$ components of the window are inputs to the deep learning model while the subsequent $\qOut$ components constitute the target output sequence for the model.}
\label{fig:construct_dataset}
\end{figure}

\textbf{Training and validation datasets}. Given the realizations at $\nTimesObs$ time stamps $\bfx(t_1),\dots,\bfx(t_{\nTimesObs})$, the training and validation datasets are partitioned as follows. Let $\eta$ be the proportion of data to be allocated to the training set. For each sequence of the given realizations, the values at the first $\lfloor \eta \nTimesObs \rfloor$ time stamps will be assigned to the training set, with the remainder forming the validation set. To construct the input-output data pairs for each dataset, we consider a sliding window of length $\qInEnc + \qOut$, applied to the time series matrix $\bfXcal$ as well as the vectors $\bfYcal$ and $\bfTcal$ of the Markov state and time sequences, as shown in Figure~\ref{fig:construct_dataset}. To illustrate, applying this procedure to $\bfXcal$, we obtain $[\bfx(t_1),\dots,\bfx(t_{\qInEnc + \qOut})]$, $[\bfx(t_2),\dots,\bfx(t_{\qInEnc + \qOut + 1})]$, \dots, $[\bfx(t_{\lfloor \eta \nTimesObs \rfloor - \qInEnc - \qOut + 1}), \dots, \bfx(t_{\lfloor \eta \nTimesObs \rfloor})]$ for training and $[\bfx(t_{\lfloor \eta \nTimesObs \rfloor + 1}), \dots, \bfx(t_{\lfloor \eta \nTimesObs \rfloor + \qInEnc + \qOut})]$, \dots, $[\bfx(t_{\nTimesObs - \qInEnc - \qOut + 1}), \dots, \bfx(t_{\nTimesObs})]$ for validation. The first $\qInEnc$ $\nSites$-dimensional vectors are inputs to the deep learning model, while the subsequent $\qOut$ vectors constitute the target output sequence for the model. For $\bfYcal$ and $\bfTcal$, the $\qInEnc + \qOut$ components all become the model inputs. There are in total $\lfloor \eta \nTimesObs \rfloor - \qInEnc - \qOut + 1$ and $(\nTimesObs - \lfloor \eta \nTimesObs \rfloor) - \qInEnc - \qOut + 1$ input-output pairs obtained from each sequence of realizations for the training and validation datasets. Note that by splitting the data first and then constructing the input-output pairs, it is ensured that there is no data leakage between training and validation datasets. We also observe that the choice of $\qInEnc$ and $\qOut$ leads to a trade-off between model accuracy and the computational efficiency of the inference procedure. Larger values of $\qInEnc$ and $\qOut$ allow for reduced iterations in the auto-regressive inference of $\bfx(t)$, leading to less computational effort. However, there are fewer input-output pairs in the training dataset which may potentially hinder the model accuracy. Conversely, smaller values of these hyperparameters increase the number of iterations in inference, while providing a greater number of input-output pairs for model training. At every training iteration, a batch of data pairs are used to update the model weights. The training and validation datasets for the deep learning model for Markov state sequence generation in Section~\ref{subsubsec:newmodel:deep_learning_model_Markov_state_generation} are constructed in a similar manner, but they only consist of the data pairs of Markov state and time sequences. 

\textbf{Loss function}. For the model in Section~\ref{subsubsec:newmodel:deep_learning_model}, we use the $L_1$ or $L_2$ loss function, also known as the mean absolute error (MAE) or mean square error (MSE), to penalize the discrepancy between the target sequence and inference of the time series by the deep learning model. For the model in Section~\ref{subsubsec:newmodel:deep_learning_model_Markov_state_generation}, we use the focal loss function \cite{focal_loss}, which is an extension of the cross entropy loss and applies a modulating term in order to focus learning on hard misclassified samples. 

\textbf{Masking}. In the training stage, \cite{attention_paper} suggests to apply triangular masks in the attention layers. This prevents each element of the sequence from attending to the elements at future times. The masking aims to mimic real scenarios in which information at future times is not available. 

\textbf{Optimizer}. We use the ADAM optimizer \cite{adam_optimizer} because of its competitive performance in deep learning applications. The learning rate is set to decay after every few epochs.

\textbf{Regularization}. Regularization serves to prevent the deep learning model from overfitting. We utilize the dropout technique in training the model weights across all layers such that a proportion of these weights are not updated in every training batch. In addition, we adopt early stopping so that  the training is automatically terminated if the validation loss is not decreasing over a certain number of epochs. This ensures that the training and validation losses are comparable which promotes the generalizability of the model to new data.

\textbf{Hyperparameter tuning}. The aforementioned architecture hyperparameters such as $\nEnc$, $\nDec$ and the hyperparameters used in the training such as the dropout rate and learning rate can be selected by hyperparameter tuning based on a test dataset. This can be achieved by grid search, greedy search, or more advanced Bayesian algorithms \cite{bayesian_optim}. 

Table~\ref{tbl: model_hyperparameters} lists the standard hyperparameters of the proposed GenFormer model.

\begin{table}
\begin{center}
\begin{tabular}{|c | c |} 
 \hline
 Notation & Description \\ 
 \hline
 $\qInEnc$ & length of the input sequence for the deep learning model  \\
 $\qOut$ & length of the output sequence for the deep learning model  \\
 $\qInDec$ & length of the start sequence for the decoder \\
 $\nClusters$ & dimension of the Markov state space used for the clustering algorithm \\
 $\dmodel$ & dimension of the hidden embedding and attention layers  \\
 $\dff$ & dimension of the hidden feed-forward network \\
 $\nHeads$ & number of heads in the attention mechanism  \\
 $\nEnc$ & number of encoder blocks in the deep learning model for inference of $\nSites$-variate processes \\
 $\nDec$ & number of decoder blocks in the deep learning model for inference of $\nSites$-variate processes \\
 $\nMarkov$ & number of decoder blocks in the deep learning model for Markov state sequence generation \\
 \hline
\end{tabular}
\caption{Standard hyperparameters of the GenFormer model.}
\label{tbl: model_hyperparameters}
\end{center}
\end{table}

\subsection{Model simulation}
\label{subsec:newmodel:model_simulation}

In the previous subsection, we discussed constructing a univariate Markov sequence via clustering and training a deep learning model with Markov state embedding based on the available observations. Here, we present the simulation methodology of GenFormer with the aim of generating new synthetic realizations of the $\nSites$-variate process over $t \in [0, \timeEndSim]$ for $\nTimesSim$ time stamps. It consists of three steps. First, a synthetic realization $\ty_1, \ldots, \ty_{\nTimesSim}$ of the univariate Markov sequence $Y_1, \ldots, Y_{\nTimesSim}$ is produced. Subsequently, the realization $\ty_1, \ldots, \ty_{\nTimesSim}$ is mapped to the preliminary synthetic realization $\tbfx(t_1), \ldots, \tbfx(t_{\nTimesSim})$ of the $\nSites$-variate stochastic process by utilizing the deep learning model with Markov state embedding. Since we want to preserve statistics of interest such as the spatial correlation matrix and marginal distributions, the final step involves model post-processing via Cholesky decomposition and the reshuffling technique.

Simulating from the univariate Markov sequence and the deep learning model requires initial data. For Markov state sequence generation, the Markov states at the first $\MarkovLag$ time stamps have to be specified. The deployment of the deep learning model requires the initial data $\bfXcal^{\text{enc}}$, $\bfYcal^{\text{enc}}$, and $\bfTcal^{\text{enc}}$ measured at the first $\qInEnc$ time points. Set $\qmax = \max(\MarkovLag, \qInEnc)$. We therefore assume that data on the $\nSites$-variate process $\tbfx(t_1), \ldots, \tbfx(t_{\qmax})$ along with the corresponding Markov state sequence $\ty_1, \ldots, \ty_{\qmax}$ at time stamps $t_1, \ldots, t_{\qmax}$ are available. To obtain such data, we randomly select a subsequence $\bfx(t_1), \ldots, \bfx(t_{\qmax})$ of length $\qmax$ and its corresponding Markov states $y_1, \ldots, y_{\qmax}$ from the given observations. 
    
    % employing Cholesky decomposition for spatial correlation matrix adjustment and a reshuffling algorithm to maintain the marginal distribution, respectively.

    % \item We present the first two steps of the simulation methodology in Section.~\ref{subsubsec:newmodel:simulation_mainsteps}, while the model post-processing procedure is elaborated in Section.~\ref{subsubsec:newmodel:simulation_postprocessing}.

\subsubsection{Simulation of $\nSites$-variate stochastic processes}
\label{subsubsec:newmodel:simulation_mainsteps}

\textbf{Simulation of a univariate Markov sequence}. Given the data $\tbfx(t_1), \ldots, \tbfx(t_{\qmax})$ with the corresponding Markov states $\ty_1, \ldots, \ty_{\qmax}$ at time stamps $t_1, \ldots, t_{\qmax}$, sampling the univariate Markov sequence is initialized using the last $\MarkovLag$ Markov states $\ty_{\qmax - \MarkovLag + 1}, \ldots, \ty_{\qmax}$. For Markov order $\MarkovLag = 1$, we utilize the estimated transition matrix to simulate a sample $\ty_{\qmax +1}$, which is repeated recursively to produce $\ty_{\qmax + 2}, \ldots, \ty_{\nTimesSim}$. For  $\MarkovLag \ge 2$, we adopt the trained deep learning model for Markov state sequence generation in Section~\ref{subsubsec:newmodel:deep_learning_model_Markov_state_generation}. In the first iteration, the model takes $\ty_{\qmax - \MarkovLag + 1}, \ldots, \ty_{\qmax}$ and $t_{\qmax - \MarkovLag + 1}, \ldots, t_{\qmax}$ as inputs to produce the sample $\ty_{\qmax +1}$ using the multinomial layer. At iteration $l$, the sequences $\ty_{\qmax - \MarkovLag + l}, \ldots, \ty_{\qmax + l-1}$ and $t_{\qmax - \MarkovLag + l}, \ldots, t_{\qmax + l-1}$ are then fed to the model to simulate $\ty_{\qmax + l}$. This process is repeated to generate $\ty_{\qmax +1}, \ty_{\qmax + 2}, \ldots, \ty_{\nTimesSim}$. 

\textbf{Inference from the deep learning model with Markov state embedding}. The second step is to infer the $\nSites$-variate stochastic process $\tbfx(t_{\qmax + 1}), \ldots, \tbfx(t_{\nTimesSim})$ corresponding to the synthetic realization $\ty_{\qmax +1}, \ldots, \ty_{\nTimesSim}$ using the deep learning model in Section~\ref{subsubsec:newmodel:deep_learning_model}. Typically, we have $\qOut \ll \nTimesSim - \qmax$ which implies that the inference needs to be performed in an auto-regressive manner. In the $l^{\text{th}}$ iteration, we infer $\tbfx(t_{\qmax + (l-1) \qOut + 1}), \ldots, \tbfx(t_{\qmax + l \qOut})$. This auto-regressive procedure is repeated $\lfloor (\nTimesSim - \qmax) / \qOut \rfloor$ times.

\subsubsection{Model post-processing}
\label{subsubsec:newmodel:simulation_postprocessing}

%Recall that we have simulated $\nSim$ samples with $\nTimes$ time steps each. It is suggested that the samples can be concatenated along the time dimension, and the reshuffling can be applied with respect to $\nTimes \times \nSim$ time stamps.

The deep learning model alone cannot fully preserve statistics of interest. We therefore introduce a model post-processing procedure. It is comprised of a transformation based on Cholesky decomposition and the reshuffling technique to correct the spatial correlation matrix and marginal distributions, respectively, of the simulated realizations. 

Let $\tbfXcal = [\tbfx(t_1), \ldots, \tbfx(t_{\nTimesSim})] \in \mathbb{R}^{\nSites \times \nTimesSim}$ be the time series matrix of preliminary synthetic realizations of $\bfX(t)$ produced by the deep learning model during inference. The spatial correlation matrix $\tbfC$ of the inference from the deep learning model can be estimated by $\tbfC = \tbfXcal  \tbfXcal^T / \nTimesSim \in \mathbb{R}^{\nSites \times \nSites}$ due to stationarity and ergodicity. Similarly, let $\bfXcal = [\bfx(t_1), \ldots, \bfx(t_{\nTimesObs})] \in \mathbb{R}^{\nSites \times \nTimesObs}$ be the matrix of observations of $\bfX(t)$. The target spatial correlation $\bfC$ of $\bfX(t)$ is approximated via $\bfC = \bfXcal  \bfXcal^T / \nTimesObs$. The accuracy of the approximation $\tbfC$ with respect to the target $\bfC$ is contingent upon the accuracy of the trained deep learning model. Consequently, we apply a transformation based on Cholesky decomposition to reduce the discrepancy between $\tbfC$ and $\bfC$.

Since the spatial correlation matrix $\tbfC$ is positive semi-definite, the Cholesky decomposition of $\tbfC$ is given by $\tbfC = \tbfL \tbfL^T$, where $\tbfL \in \R^{\nSites \times \nSites}$ is a unique lower triangular matrix \cite{cholesky}. Likewise, $\bfC = \bfL \bfL^T$ for some lower triangular matrix $\bfL$. To correct the spatial correlation matrix, we apply the transformation $\tbfU = \bfL \tbfL^T \tbfXcal$. The updated matrix $\tbfU$ has the spatial correlation matrix $\bfC$ since
\begin{align*}
    \tbfU \tbfU^T / \nTimesSim &= \bfL \tbfL^T \tbfXcal \tbfXcal^T \tbfL \bfL^T / \nTimesSim  = \bfL \tbfL^T \tbfC \tbfL \bfL^T = \bfL \textbf{I} \bfL^T  = \bfC   %\label{eq:proof_cholesky}
\end{align*}
     where $\textbf{I} \in \R^{\nSites \times \nSites}$ is the identity matrix.

The reshuffling technique discussed in Section~\ref{subsubsec:prelim:reshuffling} is then employed to rectify the marginal distributions. It is applied to the matrix $\tbfU \in \R^{\nSites \times \nTimesSim}$. We update each row $\tilde{\bfu}_i \in \R^{\nTimesSim}, i=1,\dots,\nSites,$ of $\tbfU$ by first simulating $\nTimesSim$ samples from the standard Gaussian distribution, then reshuffling these samples according to the ranks of $\tilde{\bfu}_i$, resulting in the final realization $\fsx_i \in \R^{\nTimesSim}$ for location $i$. It can be shown that the synthetic realization $\fsx_i$ at location $i$ has the standard Gaussian distribution. However, this does not guarantee that the spatial correlation matrix resulting from the transformation based on Cholesky decomposition is preserved. Nevertheless, if $\nTimesSim$ is sufficiently large, the reshuffled time series closely approximates the original, thereby minimizing the discrepancy in the spatial correlation matrix. The $j^{\text{th}}$ column of the matrix $[\fsx_1^T,\dots,\fsx_{\nSites}^T]^T \in \R^{\nSites \times \nTimesSim}$ is the final realization $\fsx(t_j)$.

\subsection{Summary of the proposed GenFormer algorithm}
\label{subsec:newmodel:simulation_algorithm_summary}

We summarize the proposed GenFormer algorithm for generating synthetic realizations of a multivariate stochastic process in Algorithm~\ref{alg:sim_alg_new}. It is comprised of the training stage and the simulation procedure. To construct the GenFormer model, the data $\bfx(t_1),\dots,\bfx(t_{\nTimesObs})$ is first transformed to have standard Gaussian marginal distributions. The $K$-means clustering algorithm in Section~\ref{subsubsec:newmodel:markov_sequence_clustering} is then employed to partition the data into $\nClusters$ clusters from which we deduce the Markov state sequence $y_1,\dots,y_{\nTimesObs}$. To fit a Markov process to the resulting state sequence, the Markov order is determined. If the order is $\MarkovLag=1$, we estimate the transition matrix whereas if $\MarkovLag\ge 2$, we train the deep learning model for Markov state sequence generation in Section~\ref{subsubsec:newmodel:deep_learning_model_Markov_state_generation}. Finally, the deep learning model with Markov state embedding is trained using  $\bfx(t_1),\dots,\bfx(t_{\nTimesObs})$ and $y_1,\dots,y_{\nTimesObs}$ following Section~\ref{subsubsec:newmodel:deep_learning_model}.

To produce synthetic realizations using GenFormer, we initialize the simulation procedure by setting $\tbfx(t_1), \ldots, \tbfx(t_{\qmax})$ and  $\ty_1, \ldots, \ty_{\qmax}$ to a randomly-chosen subsequence from the given observations. We then simulate the Markov state sequence at future times $\ty_{\qmax + 1}, \ldots, \ty_{\nTimesSim}$ using the estimated Markov transition matrix if $\MarkovLag=1$ or using the trained deep learning model for Markov state sequence generation if $\MarkovLag\ge 2$, following Section~\ref{subsubsec:newmodel:simulation_mainsteps}. Preliminary synthetic realizations $\tbfx(t_{\qmax + 1}), \ldots, \tbfx(t_{\nTimesSim})$ of the stochastic process are then obtained by applying the trained deep learning model with Markov state embedding on $\ty_{\qmax + 1}, \ldots, \ty_{\nTimesSim}$. Synthetic realizations of $\bfX(t)$ then result from post-processing the preliminary synthetic realizations to correct the spatial correlation and the marginal distributions as discussed in Section~\ref{subsubsec:newmodel:simulation_postprocessing}.

The synthetic realizations generated by GenFormer match exactly the target marginal distributions and match approximately the second-moment properties. Furthermore, the estimates based on these realizations provide satisfactory approximations to the higher-order statistical properties derived from $\bfX(t)$, even when $m$ and $\nTimesSim$ are large. We demonstrate these properties in Section~\ref{sec:numerical_examples}.

\begin{algorithm}
\caption{GenFormer model} \label{alg:sim_alg_new}
\begin{algorithmic}[1]
% \Require $\nObs$ realizations of $\bfx(t_j) = [x_1(t_j), \ldots, x_m(t_j)]^T, j = 1, \ldots, \nTimes$, transformed to have standard Gaussian marginal distributions 
\Algphase{Model construction phase}
\State Transform the data $\bfx(t_1),\dots,\bfx(t_{\nTimesObs})$ to have standard Gaussian marginal distributions
\State Partition the data into $\nClusters$ clusters via $K$-means clustering in Section~\ref{subsubsec:newmodel:markov_sequence_clustering} to obtain the Markov state sequence $y_1, \ldots, y_{\nTimesObs}$
\If{ Markov order $\MarkovLag = 1$ }
    \State Estimate the Markov transition matrix from the realizations of $y_1, \ldots, y_{\nTimesObs}$ 
\ElsIf{ Markov order $\MarkovLag \ge 2$ }
    \State Train the deep learning model for Markov state sequence generation in Section~\ref{subsubsec:newmodel:deep_learning_model_Markov_state_generation}
\EndIf
\State  Train the deep learning model with Markov state embedding in Section~\ref{subsubsec:newmodel:deep_learning_model} using the data $\bfx(t_1), \ldots, \bfx(t_{\nTimesObs})$ and $y_1, \ldots, y_{\nTimesObs}$

% Determine the lengths $\qInEnc$ and $\qInDec$ of the input sequences of encoder and decoder and the length $\qOut$ of the output sequence of the decoder.

%\State Estimate the spatial correlation matrix based on $\hat{\textbf{x}}(t_j)$ and apply Cholesky decomposition to obtain the lower triangular matrix $\hat{\textbf{L}}$
%\State Estimate the marginal distributions based on the realizations of $\hat{\textbf{x}}(t_j)$, denoted by $\hat{F}_i, i = 1, \ldots, \nSites$  \Comment{This step can be skipped if we have transformed $\hat{\textbf{x}}(t_j)$ into Gaussian space in the data pre-processing stage}

\Algphase{Model simulation phase}
\State Set $\qmax = \max(\MarkovLag, \qInEnc)$. Initialize the simulation by setting $\tbfx(t_1), \ldots, \tbfx(t_{\qmax})$ and  $\ty_1, \ldots, \ty_{\qmax}$ to a randomly chosen subsequence from the given data and its respective Markov state sequence
\If{ Markov order $\MarkovLag = 1$ }
    \State Simulate  $\ty_{\qmax + 1}, \ldots, \ty_{\nTimesSim}$ using the estimated transition matrix 
\ElsIf{ Markov order $\MarkovLag \ge 2$ }
    \State Simulate  $\ty_{\qmax + 1}, \ldots, \ty_{\nTimesSim}$ using the trained deep learning model for Markov state sequence generation according to Section~\ref{subsubsec:newmodel:simulation_mainsteps}
\EndIf
\State Apply the trained deep learning model with Markov state embedding on $\ty_{\qmax + 1}, \ldots, \ty_{\nTimesSim}$ to infer $\tbfx(t_{\qmax + 1}), \ldots, \tbfx(t_{\nTimesSim})$
\State Correct spatial correlation by applying transformation based on Cholesky decomposition in Section~\ref{subsubsec:newmodel:simulation_postprocessing} to the synthetic realizations
\State Correct the marginal distributions using the reshuffling technique in Section~\ref{subsubsec:prelim:reshuffling}
\end{algorithmic}
\end{algorithm}

\section{Numerical examples}
\label{sec:numerical_examples}

We apply the proposed GenFormer model to two examples. We first consider a synthetic dataset describing the dynamics of a $3$-variate process generated from stochastic differential equations (SDE) in Section~\ref{subsec:example:sde}. Analytical expressions for the statistical properties of the solutions to the SDE can be derived. We then consider a real dataset on wind speeds at 6 stations in Florida in Section~\ref{subsec:example:florida_wind}. The numerical results illustrate the following points:
\begin{itemize}
    \item The GenFormer model is scalable as the produced synthetic data can be used to obtain satisfactory approximations of the statistical properties of interest such as exceedance probabilities of functionals of $\bfX(t)$, even when the numbers of locations and time stamps are large. 
    \item The post-processing procedure of the GenFormer results in a model that exactly matches the target marginal distributions and reasonably approximates the spatial correlation matrix.
    
    \item For large Markov order $\MarkovLag$, the deep learning model for Markov state sequence generation is able to simulate synthetic Markov state sequences with comparable frequencies to the observed ones which is a challenge using traditional methods.
    
    \item The Markov state embedding incorporated in the deep learning model produces an accurate mapping to infer $\nSites$-variate processes from Markov state sequences.
    
\end{itemize}

In this work, we set the hyperparameters in the model architecture and training process empirically instead of employing hyperparameter tuning. This is because we already observed satisfactory performance of the trained model. However, if adequate computational resources are available, hyperparameter tuning is recommended to achieve better results. The deep learning models for Markov state sequence generation and inference of the $\nSites$-variate processes are trained based on the focal and $L_1$ losses, respectively, with batch size $128$. The ADAM optimizer \cite{adam_optimizer} is used with an initial learning rate of $10^{-4}$ which is set to decay during training. The maximum number of training epochs is 20, where the learning rate is set to be $1e^{-5}$, $5e^{-6}$, $1e^{-6}$, $5e^{-7}$ at epochs 6, 8, 10, 12, respectively. The training process is stopped early if the validation loss does not decrease over three consecutive iterations. We also employ a 5\% dropout to prevent overfitting. A Python\footnote{\href{github.com}{https://github.com/Zhaohr1990/GenFormer}} implementation of the code using PyTorch \cite{pytorch} is available. Both experiments were run on a single A100 GPU. The run time for model construction is approximately 30 minutes and the simulation of synthetic realizations can be completed within 1 minute.

%It is also worth-noting that we use separate notations, i.e., $\bfg(t)$, $\tbfg(t)$, and $\hat{\bfg}(t)$ besides $\bfx(t)$, $\tbfx(t)$, and $\hat{\bfx}(t)$, to indicate the transformed existing and synthetic realizations in the Gaussian space in the following context.  \TODO{ Think. Revisit later }

% We use $\eta = 90\%$ of the data as the training set, while the remaining is the validation set.

\subsection{Synthetic data generated from stochastic differential equations}
\label{subsec:example:sde}

\subsubsection{Problem setup}

    %\item \TODO{specific values in the training should be discussed in section 4 and not in section 3. If these values are the same for both examples, discuss them in Section 4 header, right before Section 4.1}

    %\item Move somewhere here: into a training matrix and a validation matrix based on a $90\%-10\%$ split ratio over the time dimension

    %\item Move somewhere here: where the batch size is default to be 128. 

    %\item Move somewhere here: The initial learning rate for the ADAM optimizer is set to $1e^{-4}$. In our settings, the total number of training epochs is $15$, where the learning rate is set to be $1e^{-5}$, $5e^{-6}$, $1e^{-6}$, $5e^{-7}$ at epoch 6, 8, 10, 12, respectively.

    %\item Move somewhere here: We set the dropout rate to 5\% and set the tolerance of early stopping to 3.  

    Consider the system of stochastic differential equations driven by Brownian motion with drift $a(x) = \theta(\alpha / \beta - x)$ and diffusion term $b(x) = \sqrt{2\theta x / \beta}, x \in \R,$ given by
    \begin{equation}
        d Q_i(t) = \theta \left(\frac{\alpha}{\beta} - Q_i(t) \right) dt + \sqrt{\frac{2 \theta Q_i(t)}{\beta}} dB_i(t), \quad i = 0,\dots, \nSites,\label{SDE}
    \end{equation}
    where $t \in [0, \timeEndObs)$, $Q_i(t) \in \R$, $\theta > 0, \alpha \ge 1, \beta > 0$ are prescribed coefficients, and $B_i(t), i=0,\dots, \nSites,$ are mutually independent copies of Brownian motion. It can be shown based on It\^{o}'s formula that the second-moment properties of the stationary component of $Q_i(t), i=0,\dots,\nSites,$ depend only upon the coefficient $\theta$ \cite[\MarkovLag. 435]{Grigoriu2013-yc}. More specifically, the mean and variance of the stationary process $Q_i(t)$ are $\alpha / \beta$ and $\alpha / \beta^2$, respectively. The auto-correlation function of $Q_i(t)$ has exponential decay with rate $\theta$ and is given by $\exp(-\theta \tau)$, where $\tau$ is the time lag.
    %\begin{equation}
    %    r^Q_i(\tau) = E[(Q_i(t+\tau)-\gamma^Q_1)(Q_i(t)-\gamma^Q_1)] / \gamma^Q_2  = \exp(-\theta \tau). \label{eq:gamma_2} 
    %\end{equation}
    The marginal distribution of $Q_i(t)$ follows the Gamma distribution with shape parameter $\alpha$ and rate parameter $\beta$ \cite{gamma_sde}, which can be derived through the stationary Fokker-Planck equation defined in \cite[\MarkovLag. 482]{Grigoriu2013-yc}. 

    Set $V_i(t) = Q_0(t) + Q_i(t), i = 1, \ldots, \nSites$. It can be shown that $V_i(t) \sim \text{Gamma}(2\alpha, \beta)$, i.e., $V_i(t)$ has a Gamma marginal distribution with mean $\gamma_1 = 2\alpha/\beta$ and variance $\gamma_2 = 2 \alpha / \beta^2$. The cross correlation function between $V_k(t)$ and $V_i(t)$, $1 \le k, i \le \nSites$, is given by
    \begin{equation}
        r_{ki}(\tau) =  E[(V_k(t+\tau)-\gamma_1)(V_i(t)-\gamma_1)] / \gamma_2 = \left(1 - \frac{1}{2} \mathbbm{1}(k \neq i)\right) \exp(-\theta \tau), \label{eq:gamma_cross2} 
    \end{equation}
    where $\mathbbm{1}(k \neq i)$ is the indicator function which is equal to 1 if $k \neq i$ and 0 otherwise. The normalized spatial correlation matrix of $\bfV(t)$ is given by $(r_{ki}(0))_{1 \le k, i \le \nSites}$.

    In this example, we set $\nSites = 3, \theta = 40, \alpha = \beta = 1$. A total of 1000 realizations of $V_i(t)$ are obtained by simulating sequences of $Q_i(t)$ under the Milstein scheme \cite{miltern_scheme} with duration $\timeEndObs = 0.2$ and time increment $\Delta t = 0.001$, resulting in $\nTimesObs = 200$ time steps. Since it is known that $Q_i(t) \sim \text{Gamma}(1, 1),  i = 0,1,2,3$, the numerical simulation of $Q_i(t)$ is initialized using samples from $\text{Gamma}(1, 1)$ to ensure the stationarity of $Q_i(t)$.  To pre-process the observations of $V_i(t), i = 1,2,3$, we first transform $V_i(t)$ into the standard Gaussian space via $X_i(t) = \Phi^{-1}[F(V_i(t))]$, where $F$ is the CDF of $\text{Gamma}(2, 1)$ and $\Phi^{-1}$ denotes the inverse CDF of the standard Gaussian distribution.

\subsubsection{Model considerations}

    The $K$-means clustering with $\nClusters = 300$ is applied to partition the realizations $\bfx(t) = [x_1(t), x_2(t), x_3(t)]^T$ of $\bfX(t) = [X_1(t),X_2(t),X_3(t)]^T$ across time in order to construct the state space for the discrete-time Markov process. Denote by $\{(x_1, x_2, x_3): -\infty < x_1, x_2, x_3 < \infty \}$ the sample space of $\bfX(t)$ at a fixed time $t$. We consider the tail region of $\bfX(t)$ to be $\{(x_1, x_2, x_3): \exists~i, i\in \{1,2,3\}, \text{ s.t. } x_i> q_{0.96}  \}$, where $q_{0.96} \approx 1.75$ is the $96\%$-quantile of the standard Gaussian distribution. To ensure that there is a sufficient number of cluster centroids concentrated in the tail region, we segregate the realizations into those that lie within and outside the tail region. The $K$-means clustering is then performed on each region separately, with $\nClusters = 100$ in the tail region and $\nClusters = 200$ clusters outside the tail region. We repeat the clustering process $20$ times with different starting centroids and select the clustering with the lowest within-cluster distance.

    We set the Markov order to be $\MarkovLag = 10$ and adopt a deep learning model for Markov state sequence generation with $\nMarkov = 3$ decoder blocks. The attention mechanism of the model follows the Informer \cite{informer_paper}. The dimension of the hidden attention layers is $\dmodel = 1024$ which is split among $\nHeads = 8$ heads while the dimension of the feed-forward layers is $\dff = 2048$. Since the time argument is unitless in this case, only the positional embedding is adopted as the time embedding in the embedding layer. Since the number of Markov states outside the tail region outnumbers that inside the tail region, we assign a weight of $1.3$ to the loss values induced by states in the minority class to mitigate the class imbalance issue in focal loss calculation.

    % \TODO{the fraction of states in the tail region is 1/3. Is this really imbalance? Did you not assigning weights?}

    For the deep learning model for inferring the $\nSites$-variate process, the lengths  of the input sequence of the encoder and the start sequence of the decoder are set to $\qInEnc=40$ and $\qInDec=20$, respectively. The inference length is set to be $\qOut=20$. Both the encoder and decoder have $\nEnc = \nDec = 3$ blocks while the other hyperparameters, e.g., $\nHeads$, $\dmodel$, are the same as above. We use $\eta = 0.9$ to partition the training and validation datasets. However, the approach described in Section~\ref{subsubsec:newmodel:model_training} yields insufficient data to construct the validation set. Consequently, we apply a train-validation split across realizations of sequences rather than time stamps, allocating the first $900$ realizations for training and the subsequent $100$ realizations for validation. From each sequence, $\nTimesObs - \qInEnc - \qOut + 1 = 200 - 40 - 20 + 1 = 141$ data pairs can be formed, resulting in $126900 = 900 \times 141$ pairs in the training set and $14100 = 100 \times 141$ pairs in the validation set.

    The trained GenFormer model is employed to simulate synthetic realizations on the duration of the observed data such that $\timeEndSim = \timeEndObs = 0.2$ and $\nTimesSim = \nTimesObs = 200$. For each of the $1000$ observed time series that comprise the training and validation sets, we utilize the data in the first 40 time stamps, i.e., $t\in [0,0.04)$, and their corresponding Markov state sequences, to initialize the simulation of $5$ synthetic sequences, resulting in synthetic dataset of $5000$ sequences. The trained deep learning model is then applied to generate the sequence $\tbfx(t_j), j = 41, \ldots, 200,$ from the simulated Markov states $\ty_{41}, \ldots, \ty_{200}$. Subsequently, we stack all the synthetic realizations over the time dimension and obtain a time series matrix of dimension $3 \times 10^6$. The model post processing procedures in Section~\ref{subsubsec:newmodel:simulation_postprocessing} are then applied to this concatenated matrix to obtain the final synthetic realizations $\hat{\bfx}(t)$.

\subsubsection{Results}

\begin{figure}[h]
\centering
\includegraphics[width=10cm]{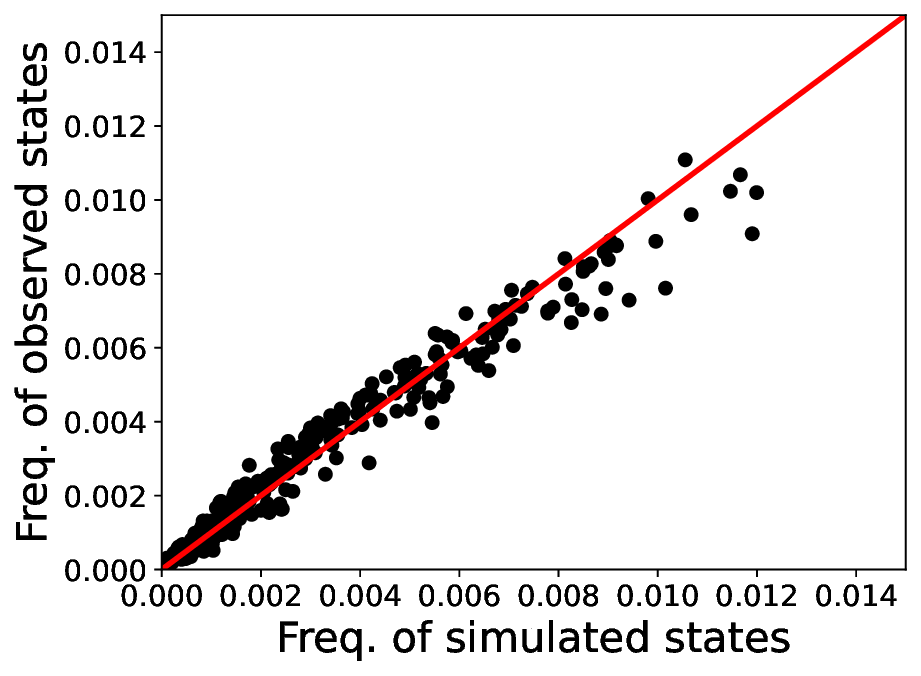}
\caption{Scatter plot of the normalized frequencies of Markov states in the observed and simulated sequences. Generating Markov state sequences by estimating the transition matrix from data is computationally challenging for large Markov order $\MarkovLag$. This example shows that for large $\MarkovLag$, the trained deep learning model for Markov state sequence generation can closely reproduce the frequencies of Markov states in the observed Markov state sequence data.}
\label{fig:state_comp_toy_example}
\end{figure}

We first examine the performance of the deep learning model for Markov state sequence generation. Using the observed Markov state sequences, we compute the normalized frequency of each of the 300 Markov states in the state space. We repeat the same procedure using the simulated Markov state sequences. In Figure~\ref{fig:state_comp_toy_example}, we show a scatter plot of the normalized frequency for each Markov state obtained from the observed sequences ($x$-axis) and the simulated sequences ($y$-axis). The approximate alignment of the scatter points along the diagonal line (red line) indicates that the frequencies of the Markov states in the observed and simulated sequences are similar. 

\begin{figure}[h]
\centering
\begin{subfigure}{.31\textwidth}
    \centering
    \includegraphics[width=.95\linewidth]{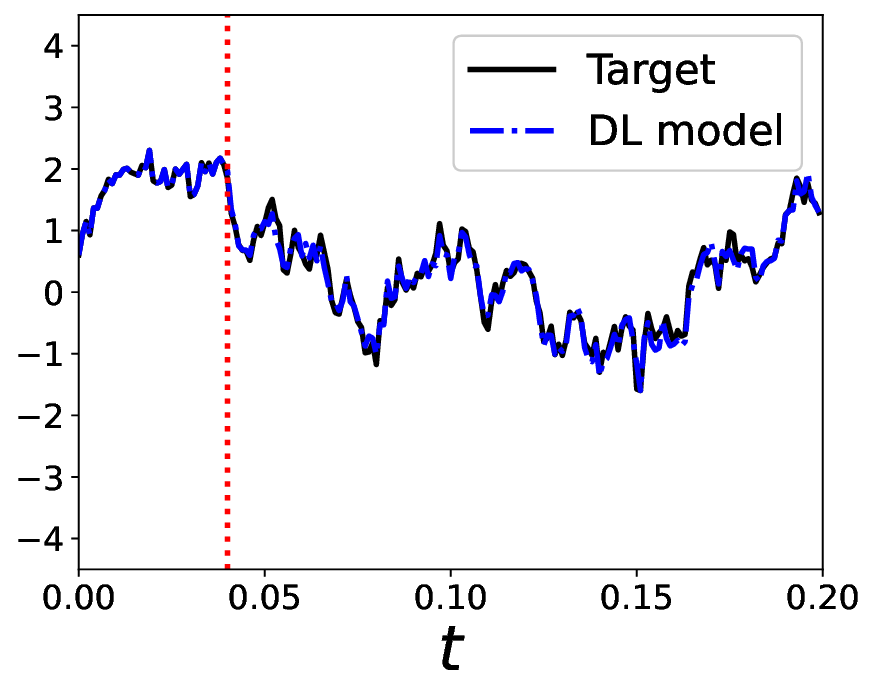} 
    \caption{$X_1(t)$}
\end{subfigure}
\begin{subfigure}{.31\textwidth}
    \centering
    \includegraphics[width=.95\linewidth]{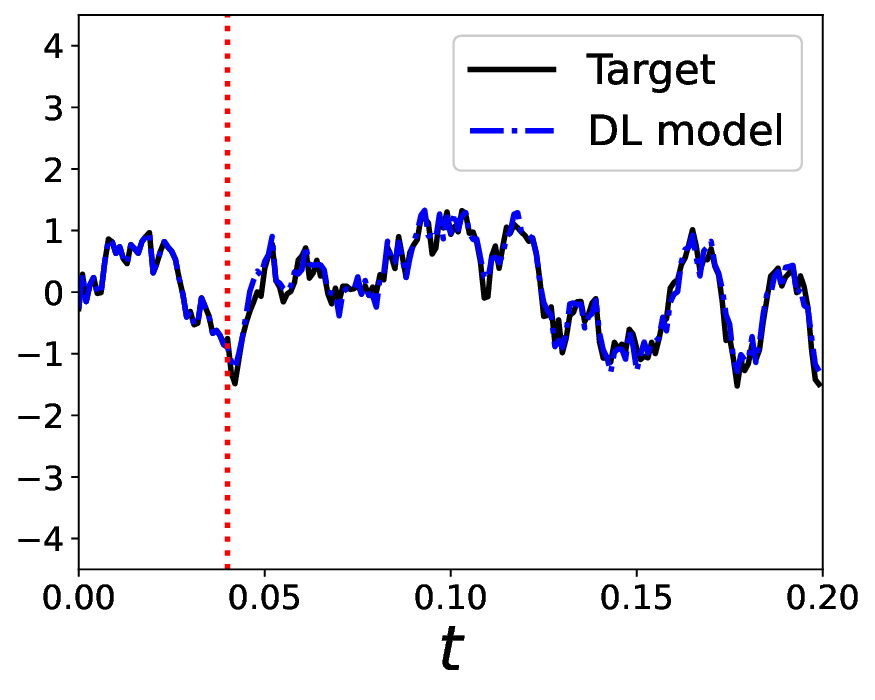}  
    \caption{$X_2(t)$}
\end{subfigure}
\begin{subfigure}{.31\textwidth}
    \centering
    \includegraphics[width=.95\linewidth]{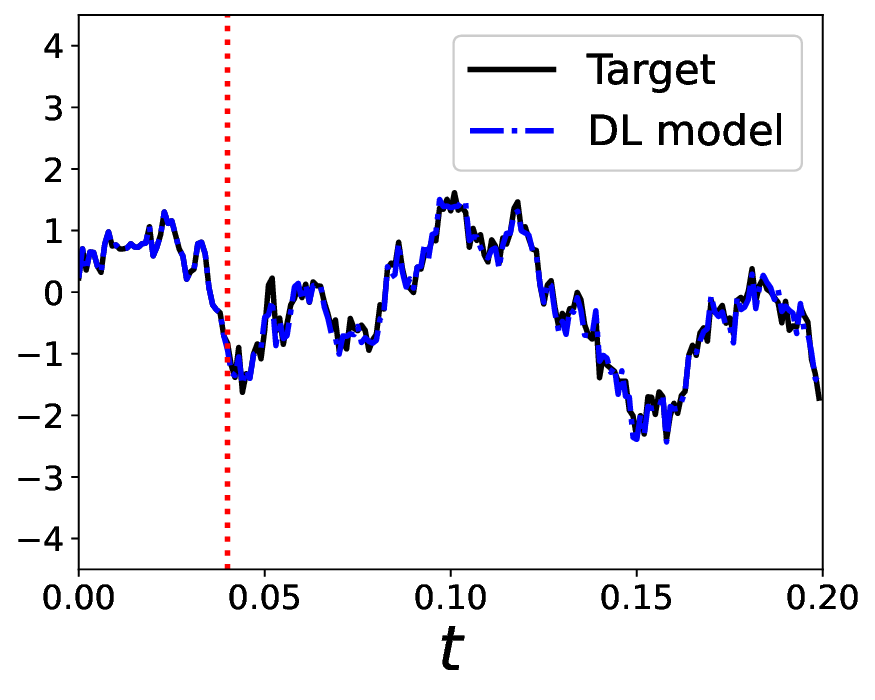}  
    \caption{$X_3(t)$}
\end{subfigure}
\caption{Target vs. synthetic time series produced by the deep learning model for inference of $\nSites$-variate processes. The transformer-based model produces accurate inference of the target based on the same Markov state sequence.}
\label{fig:ts_dl_toy_example}
\end{figure}

We then evaluate the accuracy of the deep learning model for the mapping from the Markov state $y_j$ to the time series $\bfx(t_j)$. The values of the $L_1$ losses on the training and validation datasets are $0.1145$ and $0.1199$, respectively. We visually assess the accuracy of the trained deep learning model by comparing a single observation trajectory $\bfx(t_1), \dots, \bfx(t_{200})$ with its inferred one $\tbfx(t_1), \dots, \tbfx(t_{200})$ based on the same Markov state sequence $y_1, \ldots, y_{200}$. The target trajectory is randomly selected from the validation set while the synthetic time series is produced with an initialization using the data $\bfx(t_1),\dots,\bfx(t_{40})$ and $y_1, \dots, y_{40}$ since $\qInEnc = 40$. As $\qOut = 20$, the inference is performed in an autoregressive manner comprised of 8 iterations based on the remaining sequence $y_{41}, \dots, y_{200}$. Figure~\ref{fig:ts_dl_toy_example} compares the target $\bfx(t_j), j=1,\dots,200$, and the synthetic time series $\tbfx(t_j),j=1,\dots,200$. Data to the left of the dotted red line were used to initialize the model. It can be seen that the synthetic time series accurately approximates the target.

We now examine the performance of the proposed GenFormer model in capturing desired statistical properties of the observed time series data. Figure~\ref{fig:spatial_covariance_toy_example} displays various approximations to the target spatial correlation matrix $\bfC$ in Figure~\ref{fig:spatial_covariance_toy_example}(a), estimated from the 1000 given realizations of $\bfx(t)$ in the training and validation set. The estimate in Figure~\ref{fig:spatial_covariance_toy_example}(b) is obtained using the 5000 synthetic realizations produced by the trained deep learning model in Section~\ref{subsubsec:newmodel:deep_learning_model}. Figure~\ref{fig:spatial_covariance_toy_example}(c) and Figure~\ref{fig:spatial_covariance_toy_example}(d) shows estimates resulting from samples obtained with the model post-processing steps discussed in Section~\ref{subsubsec:newmodel:simulation_postprocessing}. More specifically, Figure~\ref{fig:spatial_covariance_toy_example}(c) is based on the samples obtained after applying the transformation based on Cholesky decomposition while Figure~\ref{fig:spatial_covariance_toy_example}(d) is the approximation by the GenFormer model which subsequently applies the reshuffling procedure. For reference, Figure~\ref{fig:spatial_covariance_toy_example}(e) provides the analytical correlation matrix of $\bfV(t)$, derived from \eqref{eq:gamma_cross2}. According to \cite{Grigoriu2013-yc}, $\bfX(t)$ and $\bfV(t)$ possess roughly the same spatial correlations.

We compute the relative error between the target $\bfC$ and an approximation  $\tbfC$ via 
\begin{align}\label{eq:relErrorMatrix}
\frac{\|\bfC - \tbfC\|_F}{\|\bfC\|_F},
\end{align}
where $\| \cdot \|_F$ is the Frobenius norm. The relative errors between the matrices in (a) and (b), (a) and (c), (a) and (d) are given by 0.0411, 0.0008, 0.0045. Notice that the estimate using the samples obtained by applying a transformation based on Cholesky decomposition is able to match the target while the estimate using the samples obtained from the reshuffling procedure only induces minimal deviation. Consequently, the proposed GenFormer model produces an estimate of the spatial correlation that closely approximates the Monte Carlo estimate computed from the given realizations.

\begin{figure}[h!]
\noindent\begin{subfigure}[b]{0.32\textwidth}
    \begin{align*}
     \begin{bmatrix}
        1 & 0.47 & 0.48 \\
        0.47 & 1 & 0.45 \\
        0.48 & 0.45 & 1 
        \end{bmatrix}
    \end{align*}
    \caption{Target matrix from observations $\bfx(t)$}
\end{subfigure}
\noindent\begin{subfigure}[b]{0.3\textwidth}
    \begin{align*}
     \begin{bmatrix}
        0.98 & 0.46 & 0.50 \\
        0.46 & 0.94 & 0.44 \\
        0.50 & 0.44 & 0.96 
        \end{bmatrix}
    \end{align*}
    \caption{deep learning model estimate}
\end{subfigure} 
\noindent\begin{subfigure}[b]{0.38\textwidth}
    \begin{align*}
     \begin{bmatrix}
        1 & 0.47 & 0.48 \\
        0.47 & 1 & 0.45 \\
        0.48 & 0.45 & 1 
        \end{bmatrix}
    \end{align*}
    \caption{Cholesky-based transformation estimate}
\end{subfigure} \\
\noindent\begin{subfigure}[b]{0.5\textwidth}
    \begin{align*}
     \begin{bmatrix}
        1 & 0.47 & 0.47 \\
        0.47 & 1 & 0.45 \\
        0.47 & 0.45 & 1 
        \end{bmatrix}
    \end{align*}
    \caption{GenFormer model estimate}
\end{subfigure}%
\noindent\begin{subfigure}[b]{0.5\textwidth}
    \begin{align*}
     \begin{bmatrix}
        1 & 0.5 & 0.5 \\
        0.5 & 1 & 0.5 \\
        0.5 & 0.5 & 1 
        \end{bmatrix}
    \end{align*}
\caption{Analytical correlation matrix of $\bfV(t)$}
\end{subfigure}%
\caption{Target spatial correlation matrix of $\bfX(t)$ (a), various approximations (b), (c), (d), and analytical spatial correlation matrix of $\bfV(t)$ (e). The estimate produced by the GenFormer model has relative error that is 9 times more accurate than the estimate obtained by the deep learning model alone without the post-processing steps in this example. This highlights the need for the post-processing procedure as a supplement to the deep learning model in order to capture key statistical properties such as the spatial correlation matrix.}
\label{fig:spatial_covariance_toy_example}
\end{figure}

\begin{figure}[h]
\centering
\begin{subfigure}{.31\textwidth}
    \centering
    \includegraphics[width=.95\linewidth]{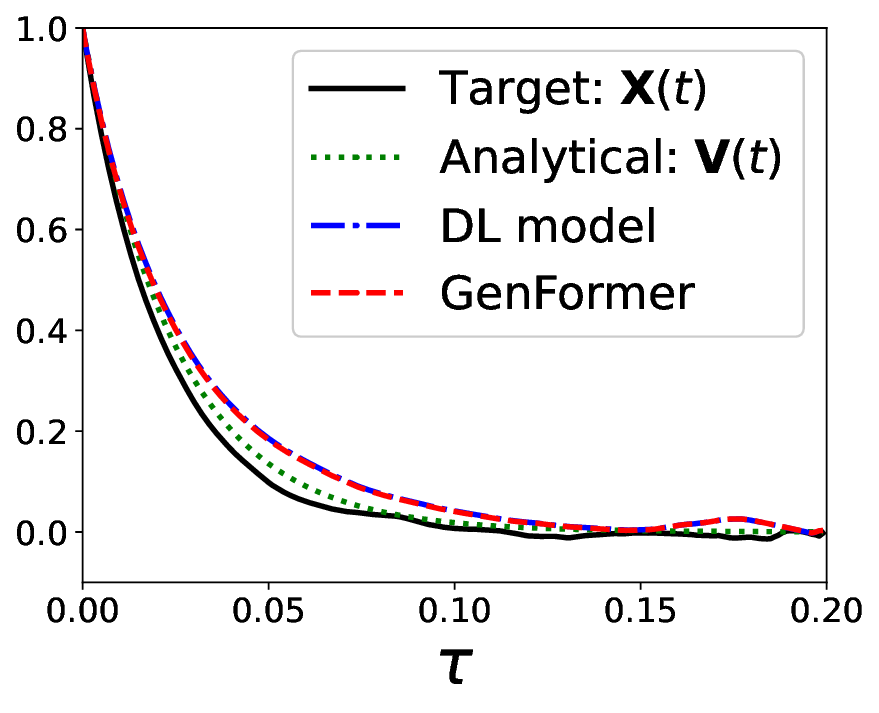} 
    \caption{$X_1(t)$}
\end{subfigure}
\begin{subfigure}{.31\textwidth}
    \centering
    \includegraphics[width=.95\linewidth]{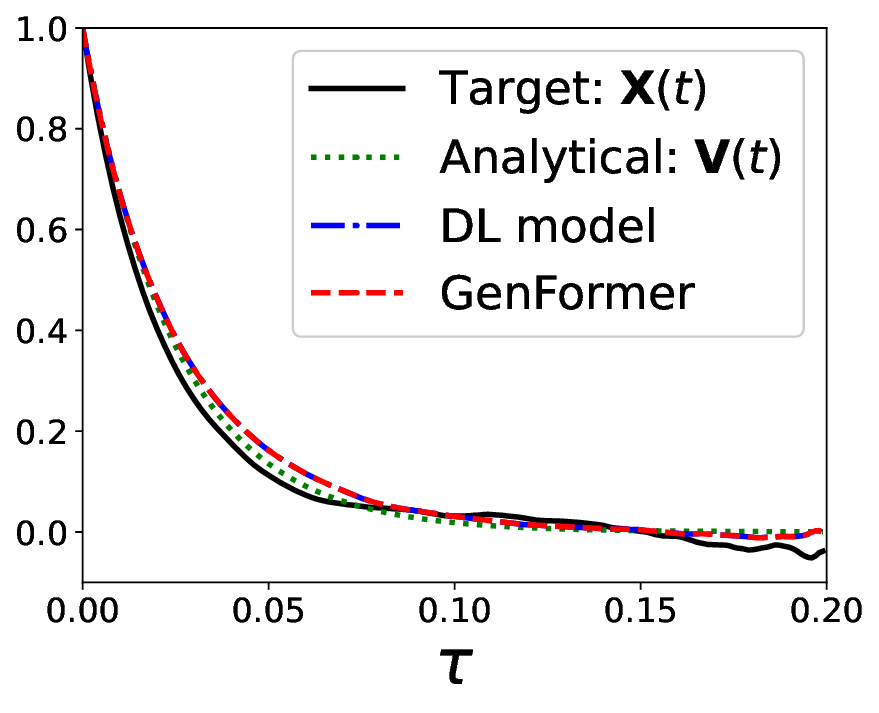}  
    \caption{$X_2(t)$}
\end{subfigure}
\begin{subfigure}{.31\textwidth}
    \centering
    \includegraphics[width=.95\linewidth]{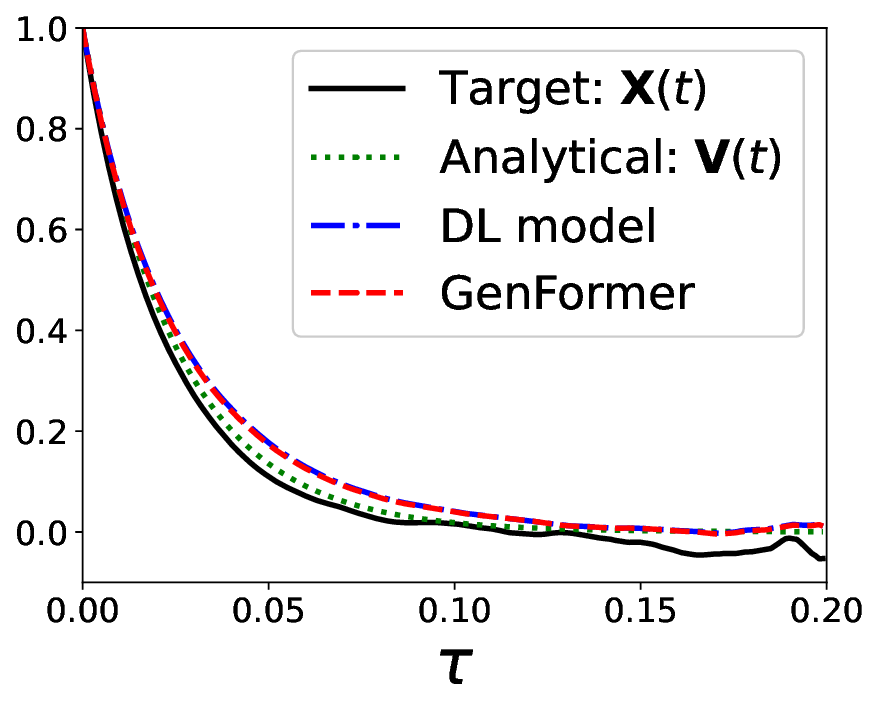}  
    \caption{$X_3(t)$}
\end{subfigure}
\caption{Auto-correlation functions of $\bfX(t)$ and various approximations. The proposed GenFormer model adequately preserves the second-moment properties of the given realizations.}
\label{fig:cor_toy_example}
\end{figure}

In Figure~\ref{fig:cor_toy_example}, we examine the auto-correlation functions of the components of $\bfX(t)$ and its various approximations. In each of the panels, the target is represented by the black solid line which is the estimate from the 1000 given realizations $\bfx(t)$. The blue dashdotted and red dashed lines are the estimates based on 5000 synthetic realizations $\tbfx(t)$ and $\hat{\bfx}(t)$ generated from the deep learning model in Section~\ref{subsubsec:newmodel:deep_learning_model} and the proposed GenFormer model, respectively. For comparison, the green dotted line is the analytical auto-correlation function of $\bfV(t)$ which closely resembles the target following \cite{Grigoriu2013-yc}. The resulting estimate from the GenFormer model provides a satisfactory approximation to the target. It can also be seen that the model post-processing procedure has negligible impact on the auto-correlation functions.

\begin{figure}[h]
\centering
\begin{subfigure}{.31\textwidth}
    \centering
    \includegraphics[width=.95\linewidth]{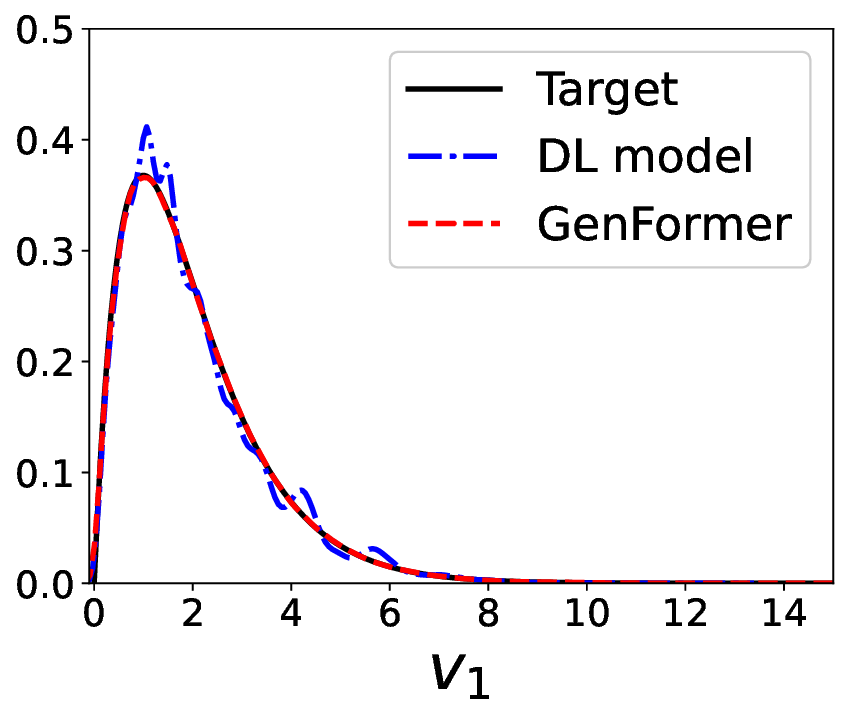} 
    \caption{Marginal density of $V_1(t)$}
\end{subfigure}
\begin{subfigure}{.31\textwidth}
    \centering
    \includegraphics[width=.95\linewidth]{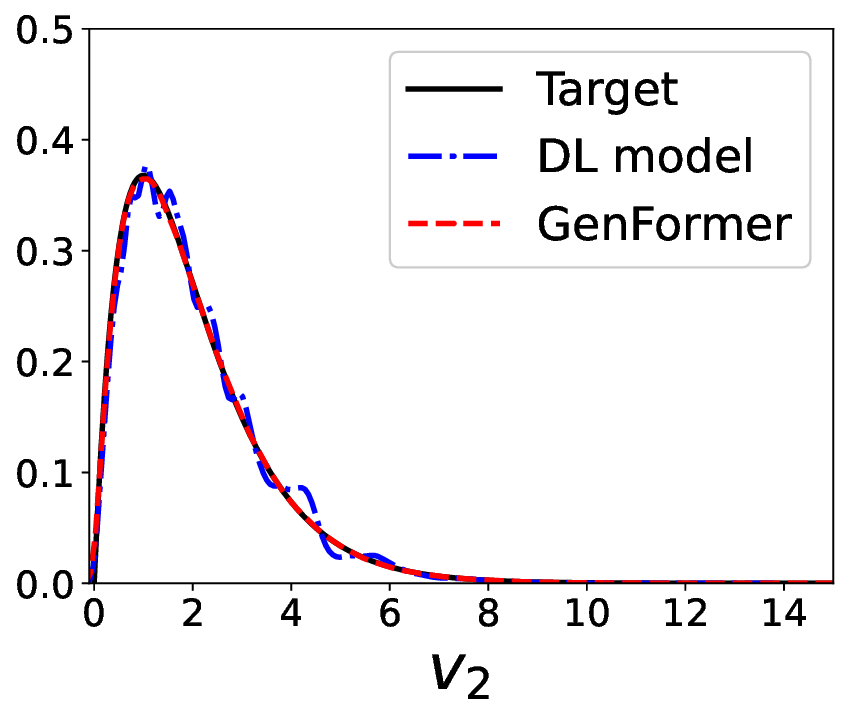}  
    \caption{Marginal density of $V_2(t)$}
\end{subfigure}
\begin{subfigure}{.31\textwidth}
    \centering
    \includegraphics[width=.95\linewidth]{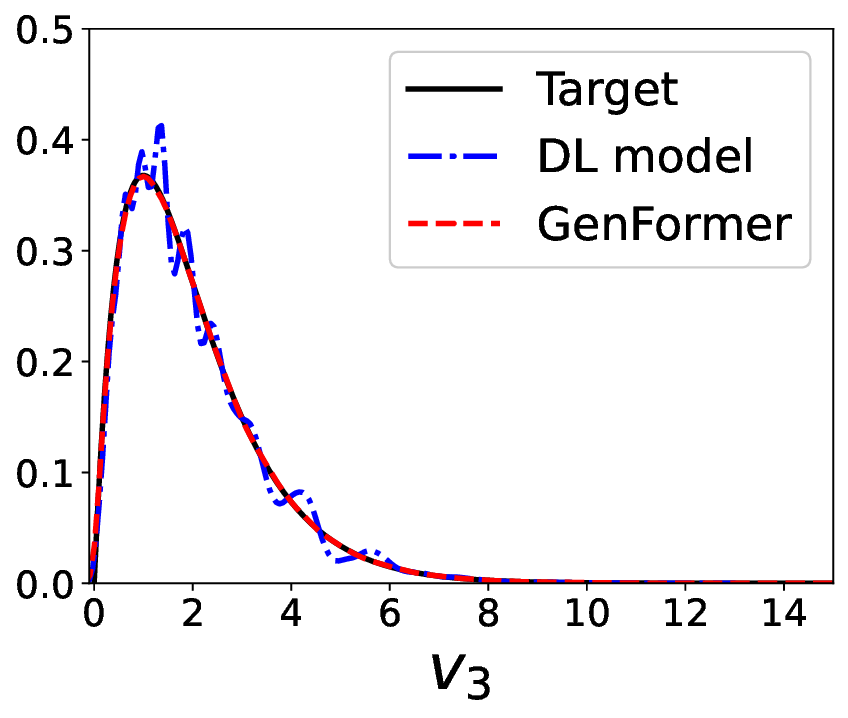}  
    \caption{Marginal density of $V_3(t)$}
\end{subfigure}
\caption{Marginal densities of $\bfV(t)$ and various approximations.  The reshuffling technique in the GenFormer model reduces the $L_1$ relative error by 1 order of magnitude in this example. This is because the target marginal distributions are directly sampled from in the reshuffling procedure.}
\label{fig:density_toy_example}
\end{figure}

In Figure~\ref{fig:density_toy_example}, we study the advantages of applying the reshuffling technique in the model post-processing procedure by visualizing estimate of the density function for each component of $\bfV(t)$. The blue dashdotted line in each panel corresponds to the inferred values of $\tbfx(t)$ from the deep learning model without the model post-processing steps applied, mapped to the original Gamma space. The red dashed line is computed from samples simulated from the GenFormer model which are subsequently transformed to have Gamma marginal distributions. The black solid line is the density of $\text{Gamma}(2, 1)$ which is the target. We calculate the $L_1$ relative errors of the density estimates with respect to the target, averaged across the 3 dimensions. The error of the density estimate computed from samples without post-processing applied is 0.1173. In contrast, the error of the density estimate resulting from samples simulated from the GenFormer model is 0.0194. This demonstrates that the GenFormer model decreases the error in the approximation of the marginal distributions by 1 order of magnitude in this example. 

Finally, we consider a downstream application of the GenFormer model in risk management. A metric of interest is the exceedance probability at a specified time $t$ defined as $p(s) = P(S(t) > s)$, where $S(t) = \sum_{i=1}^\nSites V_i(t)$. A commonly-used model for computing such probability is the translation process \cite{translation_paper_mix}, which is computationally feasible in high dimensions and serves as our baseline model for comparison. A translation process $\bfV_T(t) = [V_{T,1}(t),\dots,V_{T,\nSites}(t)]^T$ is a nonlinear memoryless transformation of a standard Gaussian process whose $i^{\text{th}}$ component is expressed as $V_{T, i}(t) = F_i^{-1}[\Phi(X^*_i(t))]$, where $F_i$ is the marginal distribution of $V_i(t)$ and $\bfX^*(t) = [X^*_1(t), \ldots, X^*_\nSites(t)]^T$ is the $\nSites$-variate Gaussian process which has the same second-moment properties (i.e., spatial correlations, auto-correlation functions, etc.) as $\bfX(t)$. However, in general, the statistical properties of $\bfX^*(t)$ and $\bfX(t)$ differ beyond the second moment. To generate synthetic realizations of $\bfV_T(t)$, the second-moment properties of $\bfX(t)$ are first estimated from the Gaussian-transformed observations of $\bfV(t)$. Subsequently, samples of $\bfX^*(t)$ are generated from a multivariate Gaussian distribution with the aforementioned second-moment properties. The mapping $F_i^{-1}[\Phi(X^*_i(t))]$ is then applied to each component of these samples to obtain samples of $\bfV_T(t)$.

\begin{figure}[h]
\centering
\includegraphics[width=9cm]{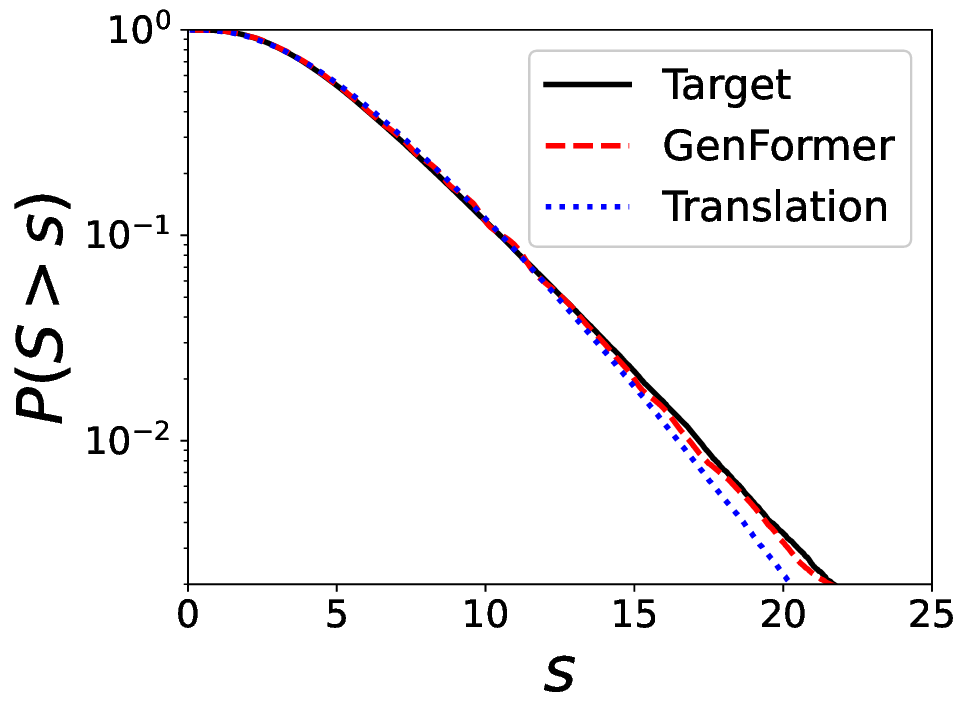}
\caption{Exceedance probability of $S(t)$. The relative error in the return period attained by the proposed GenFormer model is approximately an order of magnitude lower  than that of the translation model. The GenFormer model can capture higher-order statistical properties of $\bfX(t)$ beyond the second moment in this example.}
\label{fig:ep_toy_example}
\end{figure}

In this example, $F_i\sim \text{Gamma}(2, 1), i=1,2,3$. In Figure~\ref{fig:ep_toy_example}, we plot the exceedance probability $p(s)$ for $s \in [0,25]$ estimated using the given observations (black solid line), and the synthetic realizations produced by the translation model (blue dotted line) and the GenFormer model (red dashed line). In risk management, the inverse of the exceedance probability is the return period which quantifies the average time interval between events $\{S(t) > s\}$. The $L_1$ relative errors based on the return periods obtained from the translation model and the GenFormer model are $0.3825$ and $0.0680$, respectively. We notice that the exceedance probability curve due to the proposed GenFormer closely follows the target while the curve produced by the translation model deviates from the target as $s$ increases, resulting in underestimation of the exceedance probability and a significant error in estimating the return period for large values of $s$. The discrepancy between the two approximations to the target is due to the fact that the proposed GenFormer model is able to capture the higher-order statistical properties of $\bfX(t)$ beyond the second moment. 
    
\subsection{Simulation of station-wise wind speeds in Florida}
\label{subsec:example:florida_wind}

\subsubsection{Problem setup}

We apply the proposed GenFormer model to the hourly station-wise wind speed data\footnote{\href{https://observablehq.com/@observablehq/noaa-weather-data-by-major-u-s-city}{https://observablehq.com/@observablehq/noaa-weather-data-by-major-u-s-city}} from the National Oceanic and Atmospheric Administration (NOAA). We select 6 weather stations around Coral Gables, Florida, an area that is frequently affected by hurricanes and high wind speeds. A detailed list of station information can be found in Table.~\ref{tbl: florida_stations}. The hourly wind speed data ranging from January 1, 2006 to December 31, 2021 is collected which amounts to $\nTimesObs = 140256$ data points per station over $\timeEndObs = 5844$ days (equivalent to $16$ years). 

\begin{table}
\begin{center}
\begin{tabular}{|c | c | c | c | c |} 
 \hline
 Station ID & Station Name & State & Latitude & Longitude \\ 
 \hline
 747830 & FT LAUD/HOLLYWOOD INTL APT & FL, US & 26.079 & -80.162 \\ 
 722037 & NORTH PERRY AIRPORT & FL, US & 26.000 & -80.241 \\
 722024 & OPA LOCKA AIRPORT & FL, US & 25.910 & -80.283 \\
 722020 & MIAMI INTERNATIONAL AIRPORT & FL, US & 25.788 & -80.317 \\
 722029 & KENDALL-TAMIAMI EXEC ARPT & FL, US & 25.642 & -80.435 \\
 722026 & HOMESTEAD AFB AIRPORT & FL, US & 25.483 & -80.383 \\
 \hline
\end{tabular}
\caption{Weather stations in Florida selected in this work.}
\label{tbl: florida_stations}
\end{center}
\end{table}

The wind speed data is preprocessed to remove the trend and any periodicities, rendering it stationary. Missing wind speeds are set to 0. Station-wise hourly periodicity in a day is removed by subtracting the hourly average, computed across all days. This is followed by applying a moving average to the resulting wind speed data per station with circular padding and kernel size 720 that is equivalent to a monthly average. This moving average is then subtracted from the data obtained from the previous step, resulting in stationary wind speed data. The marginal distribution per station is estimated empirically and the data is transformed to the Gaussian space.
    
    % We now estimate the marginal distribution per station empirically, and transform the data into the Gaussian space. 
    
    % The resulting data is denoted by $\textbf{g}(t_j) = [g_1(t_j), \ldots, g_6(t_j)]^T, j = 1, \ldots, 140256$. 
    
    %The stationality is validated based on the Augmented Dickey-Fuller test \TODO{Need to calculate and input the \MarkovLag-value}. 

%\begin{figure}[h]
%\centering
%\begin{subfigure}{.49\textwidth}
%    \centering
%    \includegraphics[width=1\linewidth]{figures/figures_wind/raw_w_trend_wind.png} 
%    \caption{Raw data and its trend based on moving average}
%\end{subfigure}
%\begin{subfigure}{.49\textwidth}
%    \centering
%    \includegraphics[width=1\linewidth]{figures/figures_wind/untrend_wind.png}  
%    \caption{Wind speed data with trend removed}
%\end{subfigure}
%\caption{An illustration of (a) raw wind speed data, its trend based on moving average, and (b) the resulting data with trend removed}
%\label{fig:decomposition_toy_wind}
%\end{figure}

\subsubsection{Model considerations}

As in the previous example, we designate a tail region that is defined identically as before. We then perform $K$-means clustering with $\nClusters = 100$ in the tail region and $\nClusters = 200$ outside the tail region to capture the patterns of the extremes. 

We set the Markov order to be $\MarkovLag = 36$, and use a deep learning model for Markov state sequence generation with $\nMarkov = 3$ decoder blocks. The attention mechanism utilized in this example is the Informer. The attention layers in the model have dimension $\dmodel=512$ which is divided into $\nHeads = 8$ heads while the feed-forward layers have dimension $\dff = 2048$. Given the hourly granularity of the time argument which is provided in year-month-day-hour format, the time feature embedding described in Section~\ref{subsec:prelim:deep_learning_model} is incorporated in the embedding layer. For the focal loss computation, a weight of 1.2 is applied to the Markov states in the tail region.

In this example, we aim to infer hourly wind speed data. In the architecture of the deep learning model for inference of the $\nSites$-variate process, we set $\qInEnc = 48$, $\qInDec = 48$, and $\qOut = 48$. This implies that we infer wind speeds for the next two days based on the observed wind speeds in the previous two days. Both the encoder and decoder consists of four blocks each, i.e., $\nEnc = \nDec = 4$, while $\dmodel$, $\nHeads$, and $\dff$ are identical as above. In training the deep learning models for Markov state sequence generation and the inference of the $\nSites$-variate process, the training and validation datasets are constructed following the approach in Section~\ref{subsubsec:newmodel:model_training} with $\eta = 0.9$. 

We then simulate hourly wind speeds for 1 month using the convention that a month consists of 28 days. This means that the simulation period is $\timeEndSim=28$ days which implies that the number of simulation time stamps is $\nTimesSim = 672 = 28 \times 24$. The simulation of synthetic realizations begins by extracting the observed data at $\qmax = 48$ time stamps from  each of the $192 = 16 \times 12$ sequences of monthly data over 16 years. Each initialization sequence from the observed data is used 10 times to generate 10 sequences, resulting in 1920 synthetic realizations of sequences of length 28 days. 

    %\item In practice, compared to getting $\nObs$ independent realizations, it is often the case that $\bfx(t)$ is collected over a long time duration. For example, suppose we want to simulate the potential realizations for next year. What we usually have in hand is the historical observations over the last $N_{year}$ years. Under the assumption of stationarity, we can consider $\nObs = N_{year}$, and each historical year serves as an independent realization of the data. \TODO{Only bring this up in the relevant example because it detracts from the message of this section}
    
    %\begin{itemize}
     %\item Before diving into the simulation methodology, it is also worth noting about the necessary data-preprocessing steps. First, the raw data should be imputed in case there is missing values. Second, the stationality of the time series needs to be validated, e.g., through the Augmented Dickey-Fuller test \cite{adfuller_test}. If the time series does not satisfy the stationary assumption, a common technique is to deduct the moving-average with a certain window such that the trend and seasonality of the time series can be removed \cite{autoformer_paper}.  
    %\end{itemize}

\subsubsection{Results}

\begin{figure}[h]
\centering
\includegraphics[width=10cm]{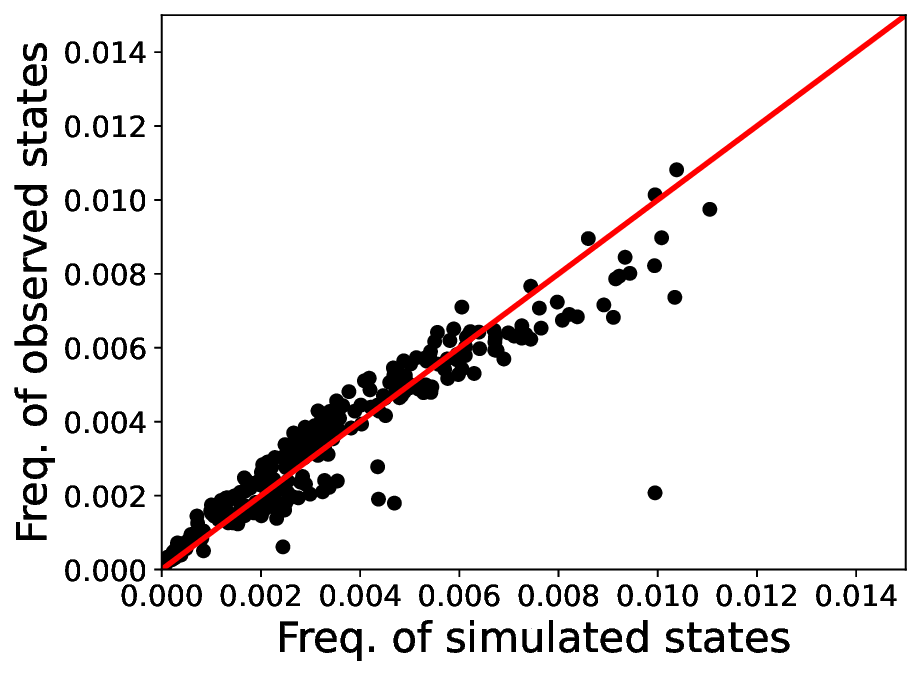}
\caption{Scatter plot of the normalized frequencies of Markov states in the observed and simulated sequences. Estimating the transition matrix for large $\MarkovLag$ is prohibitive since the transition matrix would have dimension $300^{36} \times 300$. In this example, the trained deep learning model for Markov state sequence generation  offers a computationally feasible alternative for producing synthetic Markov state sequences with the occurrence frequency of each Markov state being similar to the observed one.}
\label{fig:state_comp_toy_wind}
\end{figure}

We first evaluate the performance of the deep learning model for Markov state sequence generation in Figure~\ref{fig:state_comp_toy_wind}. A scatter plot showing the normalized frequency of each Markov state computed from the observed Markov state sequences versus the simulated Markov state sequences from the deep learning model is presented. Despite a large $\MarkovLag$ in this example, the majority of the points in the scatter plot are aligned with the red diagonal line with only a few outliers. This indicates that the trained deep learning model can closely replicate the distribution of Markov state occurrences in the observed data, even when $\MarkovLag$ is large. 

\begin{figure}[h]
\centering
\begin{subfigure}{.31\textwidth}
    \centering
    \includegraphics[width=0.95\linewidth]{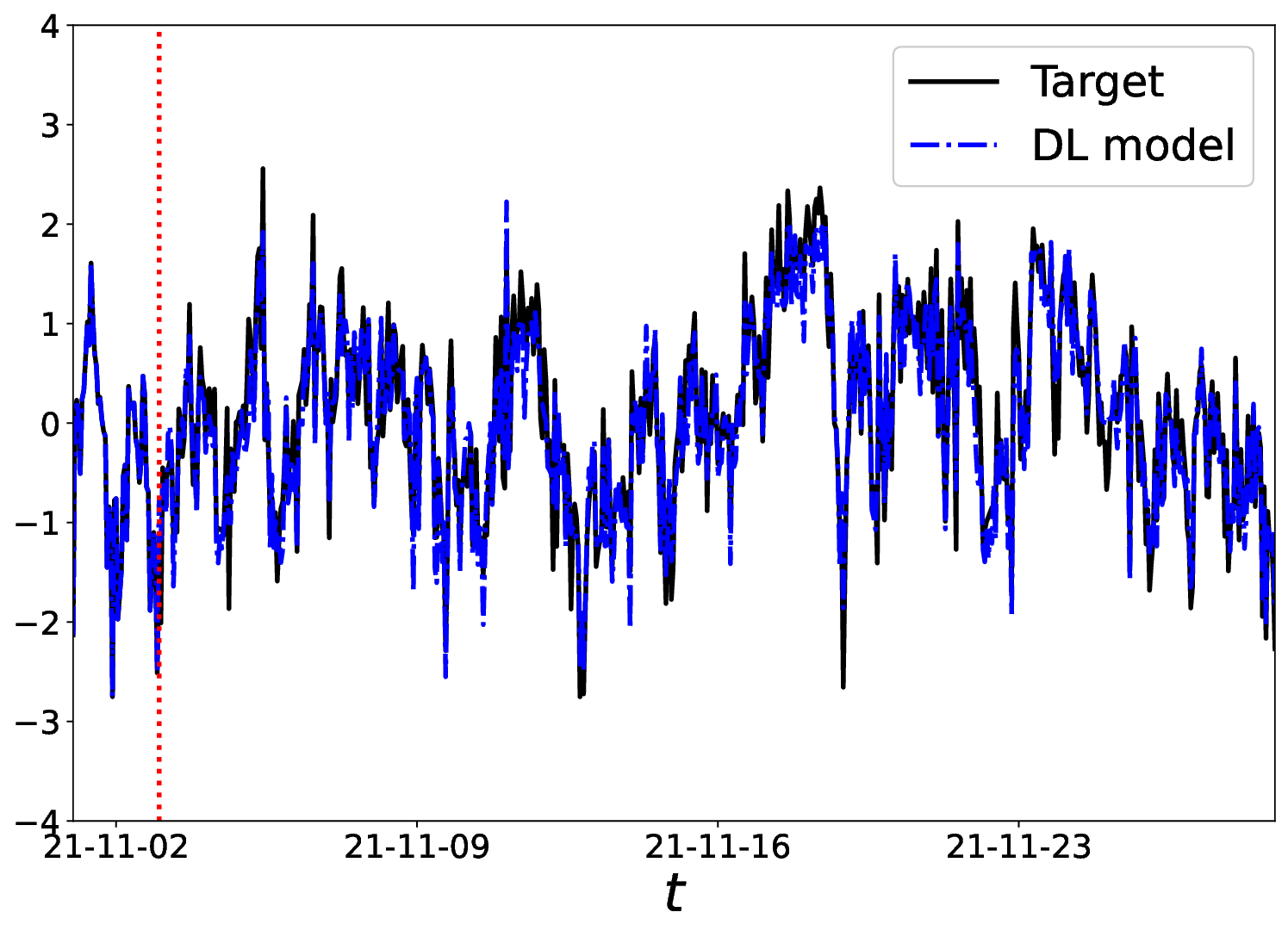} 
    \caption{$X_1(t)$}
\end{subfigure}
\begin{subfigure}{.31\textwidth}
    \centering
    \includegraphics[width=0.95\linewidth]{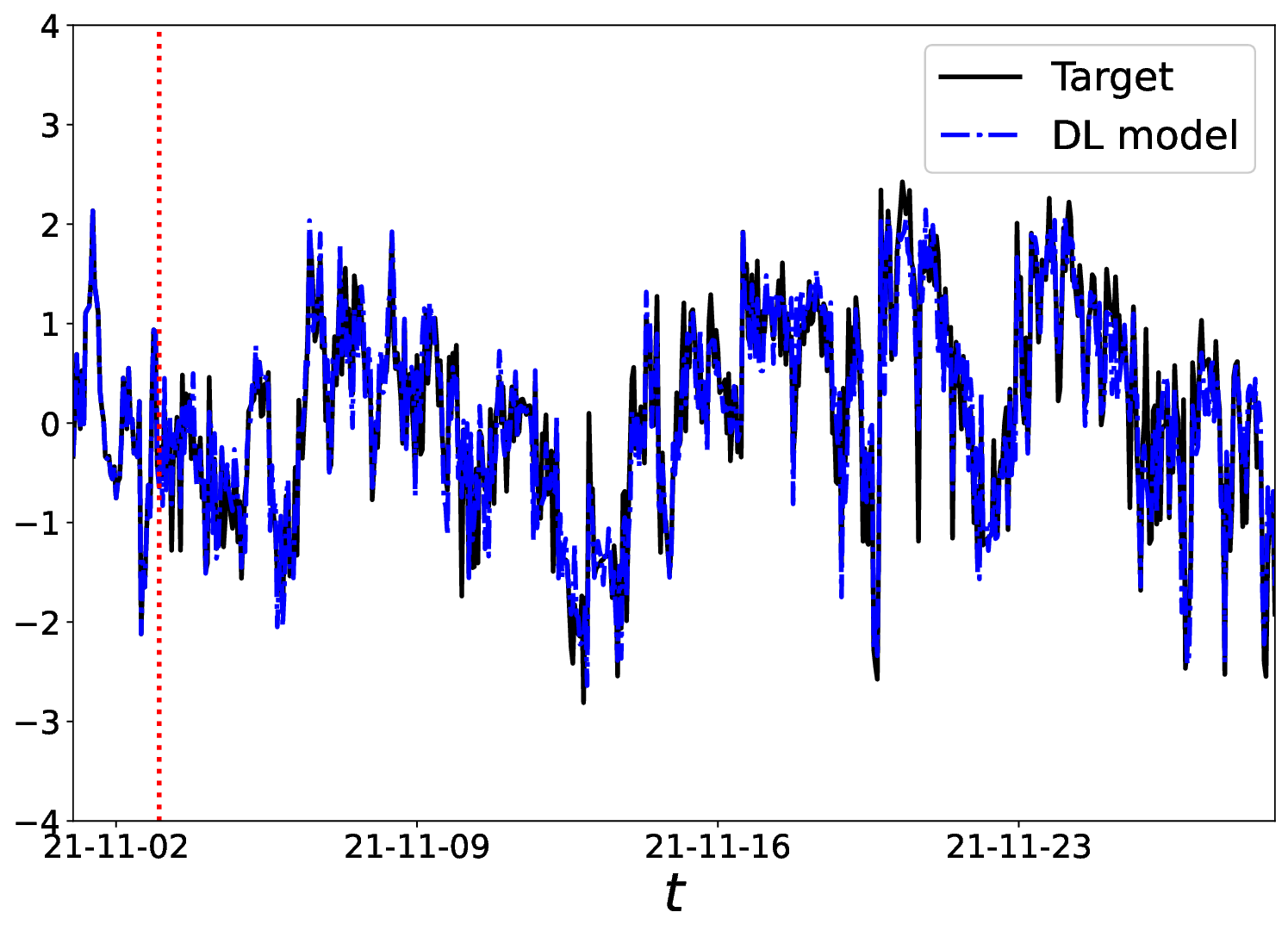}  
    \caption{$X_3(t)$}
\end{subfigure}
\begin{subfigure}{.31\textwidth}
    \centering
    \includegraphics[width=0.95\linewidth]{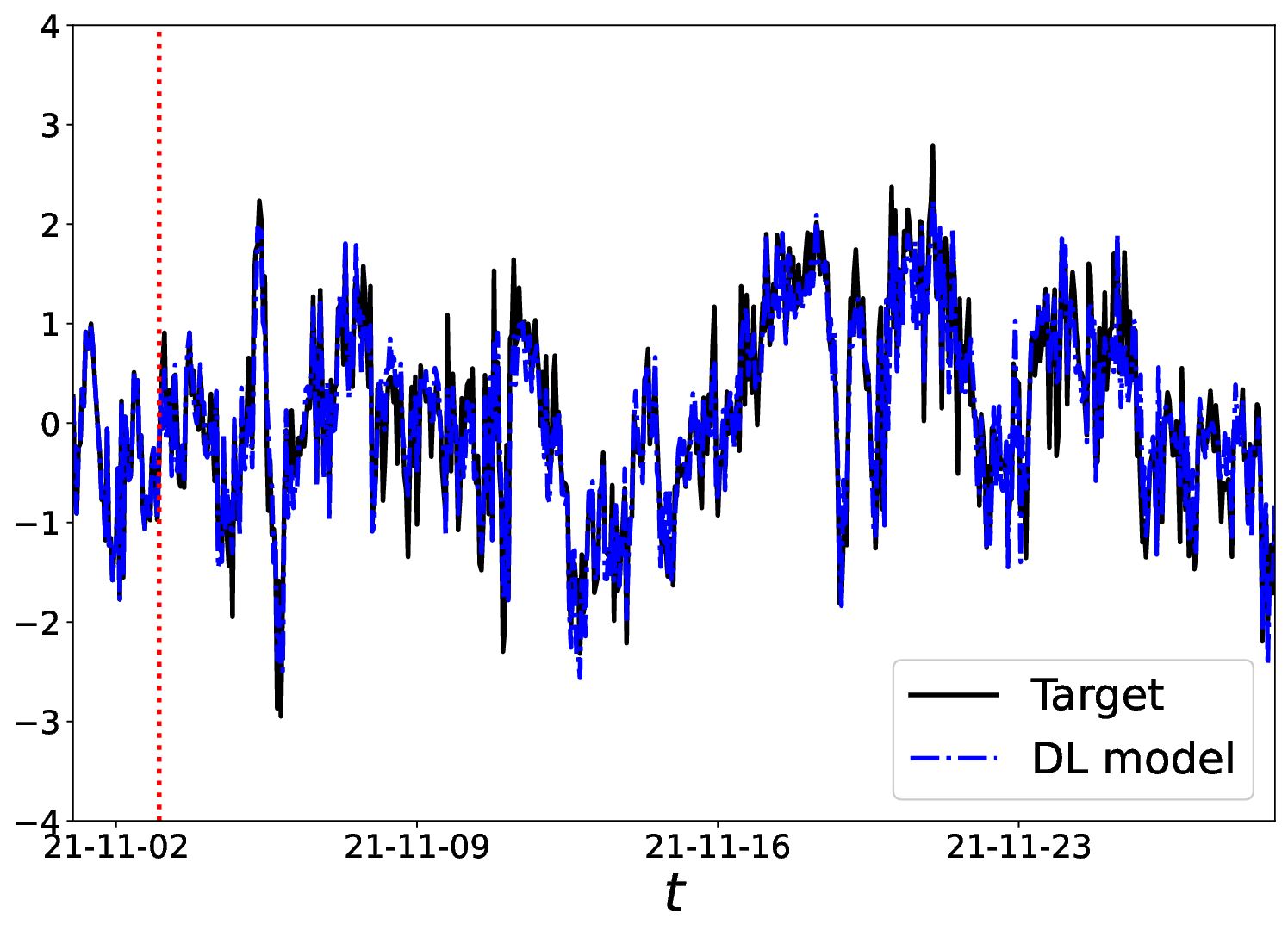}  
    \caption{$X_6(t)$}
\end{subfigure}
\caption{Target vs. synthetic time series produced by the deep learning model for inference of wind speed time series. Even though the time series data in this example is higher-dimensional and exhibits more volatility, the autoregressive inference from the Transformer-based deep learning model can still effectively approximate the target time series using a modest computational time. The inferred time series does not diverge from the target despite the long simulation horizon.}
\label{fig:ts_dl_toy_wind}
\end{figure}

The trained deep learning model for inference of the $6$-variate wind speed time series achieves $L_1$ losses of 0.2296 and 0.2504 on the training and validation sets, respectively. We visually inspect the accuracy of the trained model by utilizing it to reproduce a randomly-chosen time series in the validation set provided that the exact Markov state sequence is known, as carried out in the previous example. The inference is initialized with data from the first 2 days (48 hours), corresponding to 48 time stamps, while the inference horizon consists of the subsequent 26 days. Synthetic realizations of the wind speeds are generated using the deep learning model in an auto-regressive manner for 13 iterations since $\qOut = 2$ days. Figure~\ref{fig:ts_dl_toy_wind} plots the wind speeds simulated by the deep learning model as well as the observed wind speeds for $X_1(t),X_3(t),X_6(t)$. Data to the left of the dotted red line is used to initialize the deep learning model. Although the time series data encompasses more spatial locations and the inference is carried out over a longer simulation period with increased volatility compared to the previous example, the plots demonstrate that the trained deep learning model is still able to closely approximate the target time series.
    
\begin{figure}[h!]
\noindent\begin{subfigure}[b]{0.5\textwidth}
    \begin{align*}
        \begin{bmatrix}
            1 & 0.71 & 0.61 & 0.68 & 0.68 & 0.68 \\
            0.71 & 1 & 0.61 & 0.68 & 0.78 & 0.67 \\
            0.61 & 0.61 & 1 & 0.68 & 0.62 & 0.60 \\
            0.68 & 0.68 & 0.68 & 1 & 0.67 & 0.60 \\
            0.68 & 0.78 & 0.62 & 0.67 & 1 & 0.72 \\
            0.68 & 0.67 & 0.60 & 0.60 & 0.72 & 1 
        \end{bmatrix} 
    \end{align*}
\caption{Estimate from collected data}
\end{subfigure}%
\noindent\begin{subfigure}[b]{0.5\textwidth}
    \begin{align*}
        \begin{bmatrix}
            0.93 & 0.77 & 0.69 & 0.74 & 0.74 & 0.74 \\
            0.77 & 0.94 & 0.69 & 0.74 & 0.83 & 0.74 \\
            0.69 & 0.69 & 0.98 & 0.74 & 0.69 & 0.67 \\
            0.74 & 0.74 & 0.74 & 0.92 & 0.72 & 0.66 \\
            0.74 & 0.83 & 0.69 & 0.72 & 0.92 & 0.77 \\
            0.74 & 0.74 & 0.67 & 0.66 & 0.77 & 0.91 
        \end{bmatrix} 
    \end{align*}
    \caption{Estimate from deep learning model}
\end{subfigure} \\
\noindent\begin{subfigure}[b]{0.5\textwidth}
    \begin{align*}
        \begin{bmatrix}
            1 & 0.71 & 0.61 & 0.68 & 0.68 & 0.68 \\
            0.71 & 1 & 0.61 & 0.68 & 0.78 & 0.67 \\
            0.61 & 0.61 & 1 & 0.68 & 0.62 & 0.60 \\
            0.68 & 0.68 & 0.68 & 1 & 0.67 & 0.60 \\
            0.68 & 0.78 & 0.62 & 0.67 & 1 & 0.72 \\
            0.68 & 0.67 & 0.60 & 0.60 & 0.72 & 1 
        \end{bmatrix} 
    \end{align*}
    \caption{Estimate from Cholesky-based transformation}
\end{subfigure}%
\noindent\begin{subfigure}[b]{0.5\textwidth}
    \begin{align*}
        \begin{bmatrix}
            1 & 0.71 & 0.61 & 0.68 & 0.68 & 0.68 \\
            0.71 & 1 & 0.61 & 0.68 & 0.78 & 0.67 \\
            0.61 & 0.61 & 1 & 0.68 & 0.62 & 0.60 \\
            0.68 & 0.68 & 0.68 & 1 & 0.67 & 0.60 \\
            0.68 & 0.78 & 0.62 & 0.67 & 1 & 0.72 \\
            0.68 & 0.67 & 0.60 & 0.60 & 0.72 & 1 
        \end{bmatrix} 
    \end{align*}
    \caption{Estimate from the GenFormer model}
\end{subfigure}%
\caption{Target spatial correlation matrix of collected wind speeds (a) and various approximations (b), (c), (d).  The estimate of the spatial correlation matrix due to the GenFormer model is 12 times more accurate than the estimate produced by the deep learning model without post-processing. On the other hand, the transformation based on Cholesky decomposition preserves the spatial correlation matrix irrespective of the number of locations $\nSites$.}
\label{fig:spatial_cor_toy_wind}
\end{figure}

We now assess how well the proposed GenFormer model  captures statistical properties of interest. Figure~\ref{fig:spatial_cor_toy_wind} compares the target spatial correlation matrix (Figure~\ref{fig:spatial_cor_toy_wind} (a)) computed from the collected wind speed observations with various approximations. Figure~\ref{fig:spatial_cor_toy_wind} (b) shows the estimate computed from samples generated by the trained deep learning model. Figure~\ref{fig:spatial_cor_toy_wind} (c) is the estimate from samples to which a transformation based on Cholesky decomposition has been applied while Figure~\ref{fig:spatial_cor_toy_wind} (d) is the resulting estimate provided by the GenFormer model. The relative errors \eqref{eq:relErrorMatrix} between the matrices in (a) and (b), (a) and (c), (a) and (d) are given by 0.0862, 0.0031, and 0.0072, respectively. It can be seen that the Figure~\ref{fig:spatial_cor_toy_wind} (c) best aligns with the target since the applied transformation corrects for discrepancies in the spatial correlation. In addition, the estimate due to the GenFormer model is still comparable to the target which highlights that the reshuffling procedure in this example mostly preserves the effect of correcting the spatial correlation in the post-processing step.

\begin{figure}[h]
\centering
\begin{subfigure}{.31\textwidth}
    \centering
    \includegraphics[width=.95\linewidth]{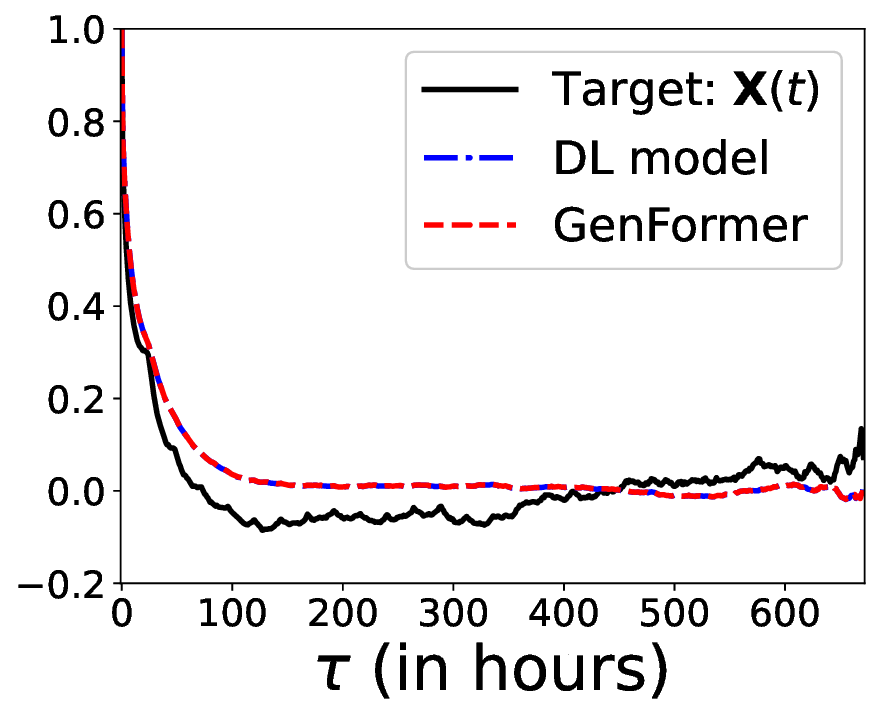} 
    \caption{$X_1(t)$}
\end{subfigure}
\begin{subfigure}{.31\textwidth}
    \centering
    \includegraphics[width=.95\linewidth]{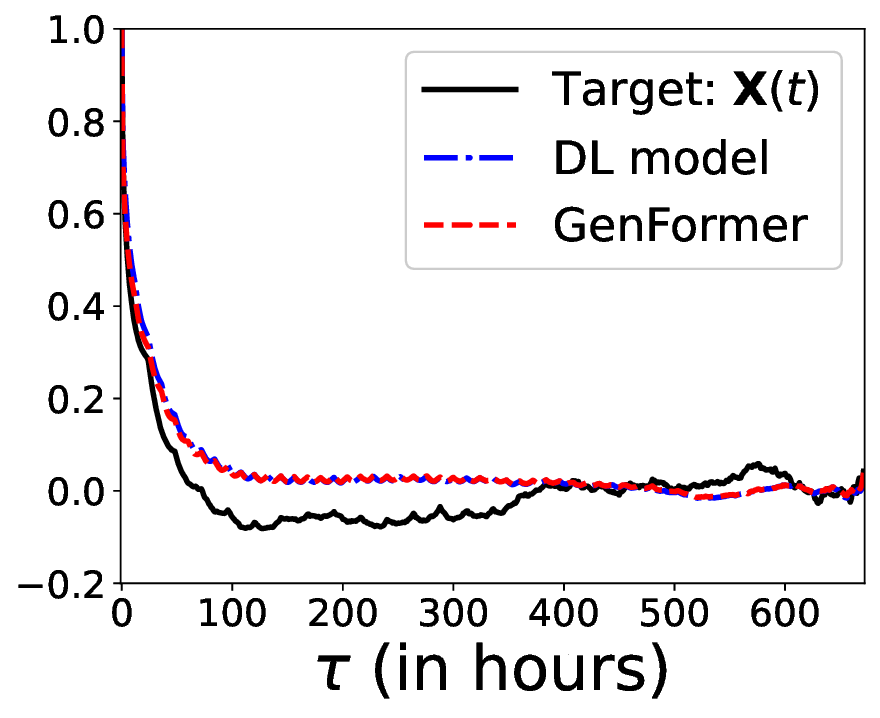}  
    \caption{$X_3(t)$}
\end{subfigure}
\begin{subfigure}{.31\textwidth}
    \centering
    \includegraphics[width=.95\linewidth]{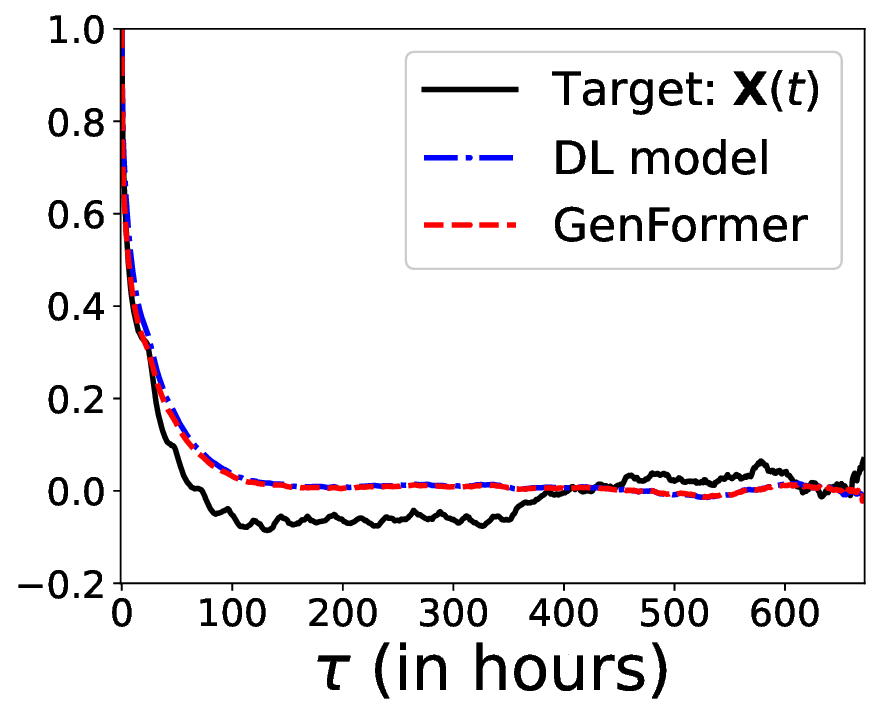}  
    \caption{$X_6(t)$}
\end{subfigure}
\caption{Auto-correlation functions of $X_1(t),X_3(t),X_6(t)$ and various approximations. The trained deep learning model provides satisfactory approximations to the auto-correlation functions with the post-processing procedure introducing only minimal and visually-indiscernible deviations.}
\label{fig:cor_toy_wind}
\end{figure}

Figure~\ref{fig:cor_toy_wind} plots the auto-correlation functions of $X_1(t)$, $X_3(t)$, and $X_6(t)$ obtained from the collected wind speed data (black solid line), samples simulated from the trained deep learning model (blue dashdotted line), and samples generated using the GenFormer model (red dashed line). The estimate due to the proposed approach offers a satisfactory approximation to the Monte Carlo estimate from the given observations.

\begin{figure}[h]
\centering
\begin{subfigure}{.31\textwidth}
    \centering
    \includegraphics[width=.95\linewidth]{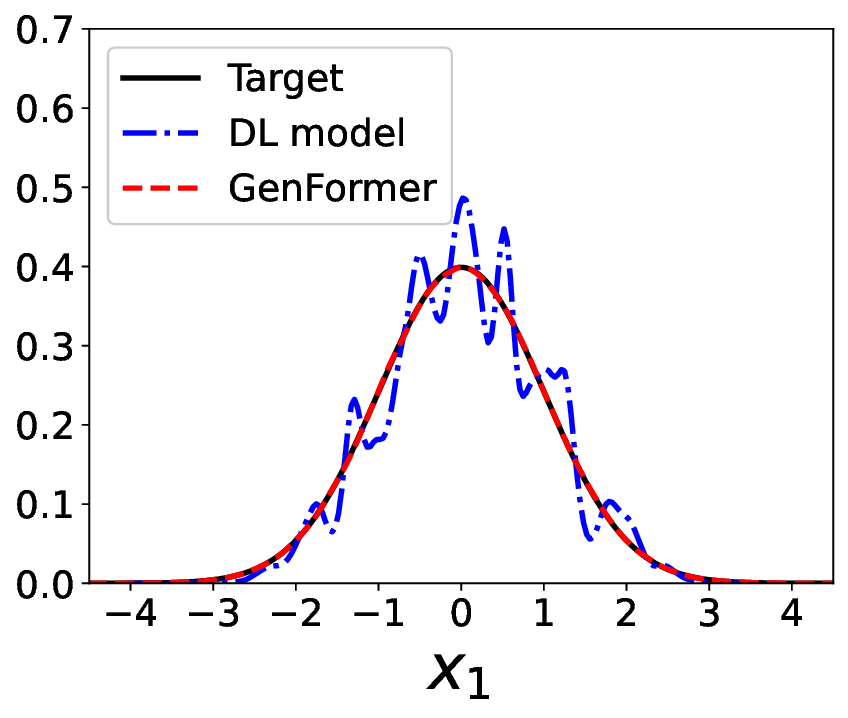} 
    \caption{Marginal density of $X_1(t)$}
\end{subfigure}
\begin{subfigure}{.31\textwidth}
    \centering
    \includegraphics[width=.95\linewidth]{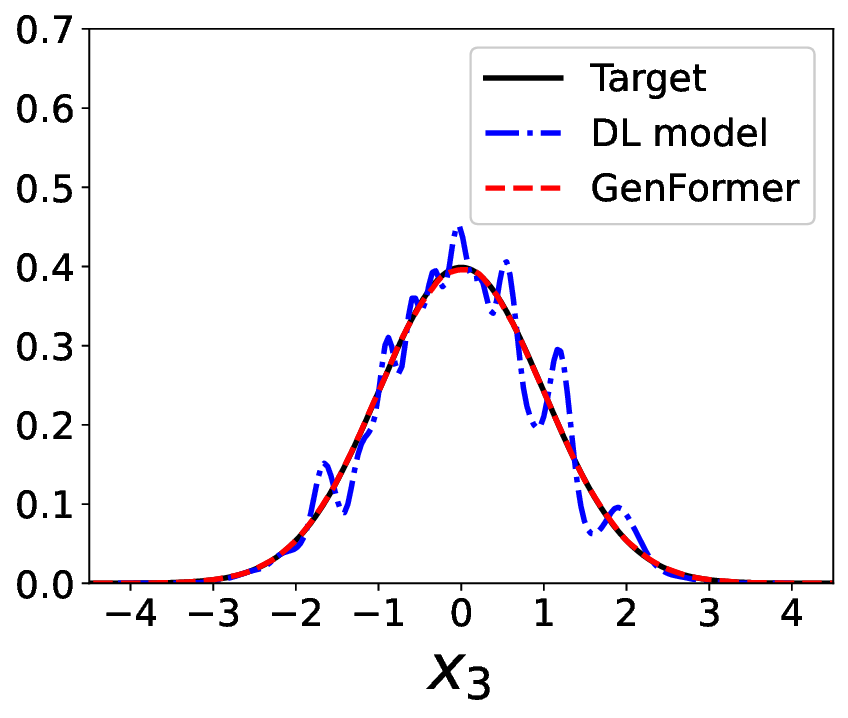}  
    \caption{Marginal density of $X_3(t)$}
\end{subfigure}
\begin{subfigure}{.31\textwidth}
    \centering
    \includegraphics[width=.95\linewidth]{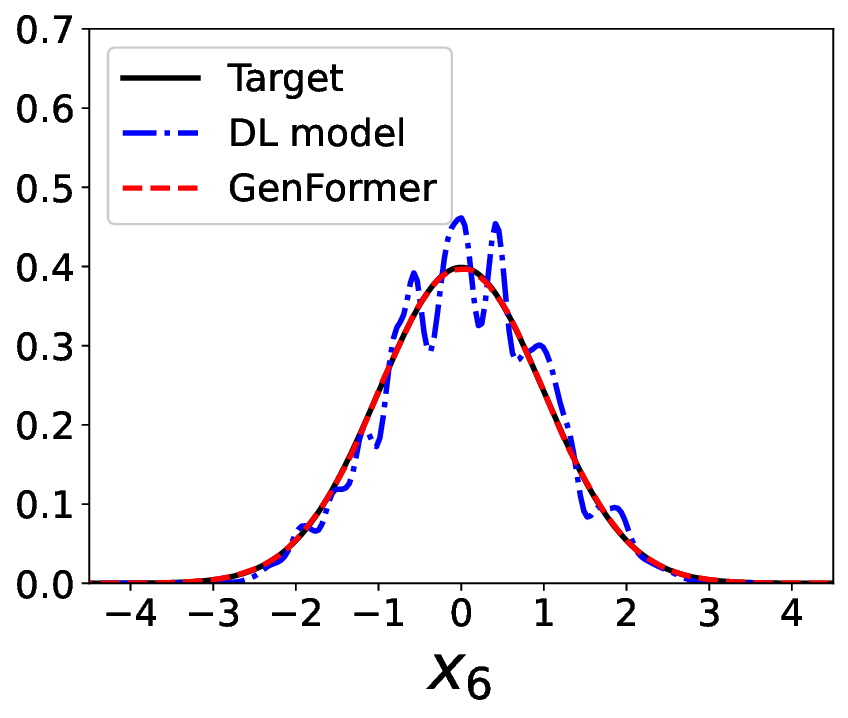}  
    \caption{Marginal density of $X_6(t)$}
\end{subfigure}
\caption{Marginal density estimates of $X_1(t), X_3(t), X_6(t)$. The model post-processing procedure in the GenFormer model, specifically the reshuffling technique, reduces the $L_1$ relative error by a factor of 11. The deep learning model alone is unable to produce samples with accurate marginal distributions because the training procedure only penalizes the discrepancy in the inferred values. }
\label{fig:density_toy_wind}
\end{figure}

In Figure~\ref{fig:density_toy_wind}, we show  estimates of the marginal densities of $X_1(t)$, $X_3(t)$, and $X_6(t)$ in the Gaussian space. The black solid line marks the target standard Gaussian density. The blue dashdotted line is the estimate computed using inferences from the deep learning model prior to the model post-processing steps with average $L_1$ relative error across the 6 spatial locations being 0.1756. The red dashed line represents the estimate due to the GenFormer model which attains an average $L_1$ relative error of 0.0157. Thus, the model post-processing procedure in the proposed GenFormer model serves to correct the marginal density of the synthetic realizations which the deep learning model alone does not account for in its training process.
    
\begin{figure}[h]
\centering
\begin{subfigure}{.31\textwidth}
    \centering
    \includegraphics[width=0.95\linewidth]{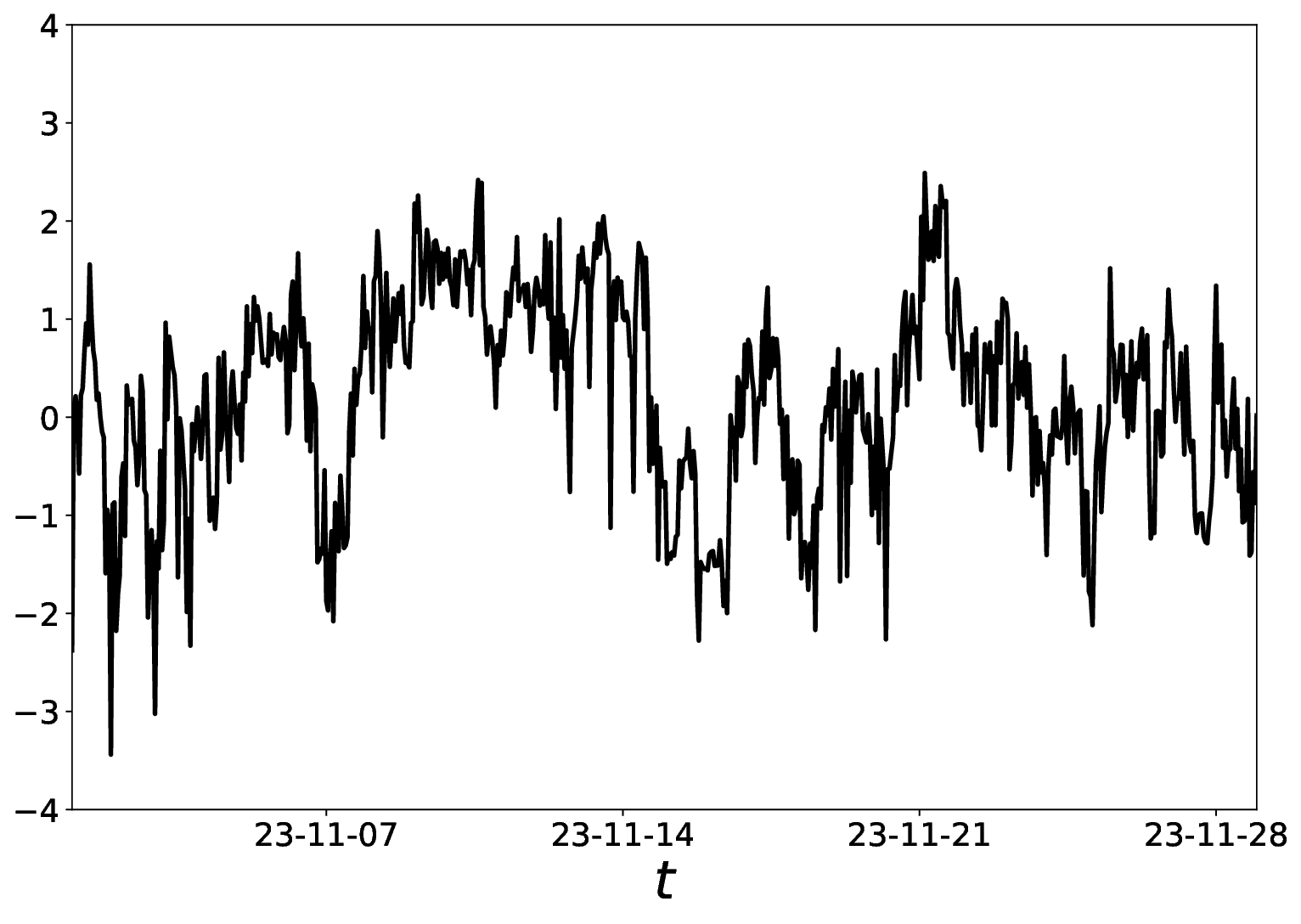} 
    \caption{Sample of $X_1(t)$}
\end{subfigure}
\begin{subfigure}{.31\textwidth}
    \centering
    \includegraphics[width=0.95\linewidth]{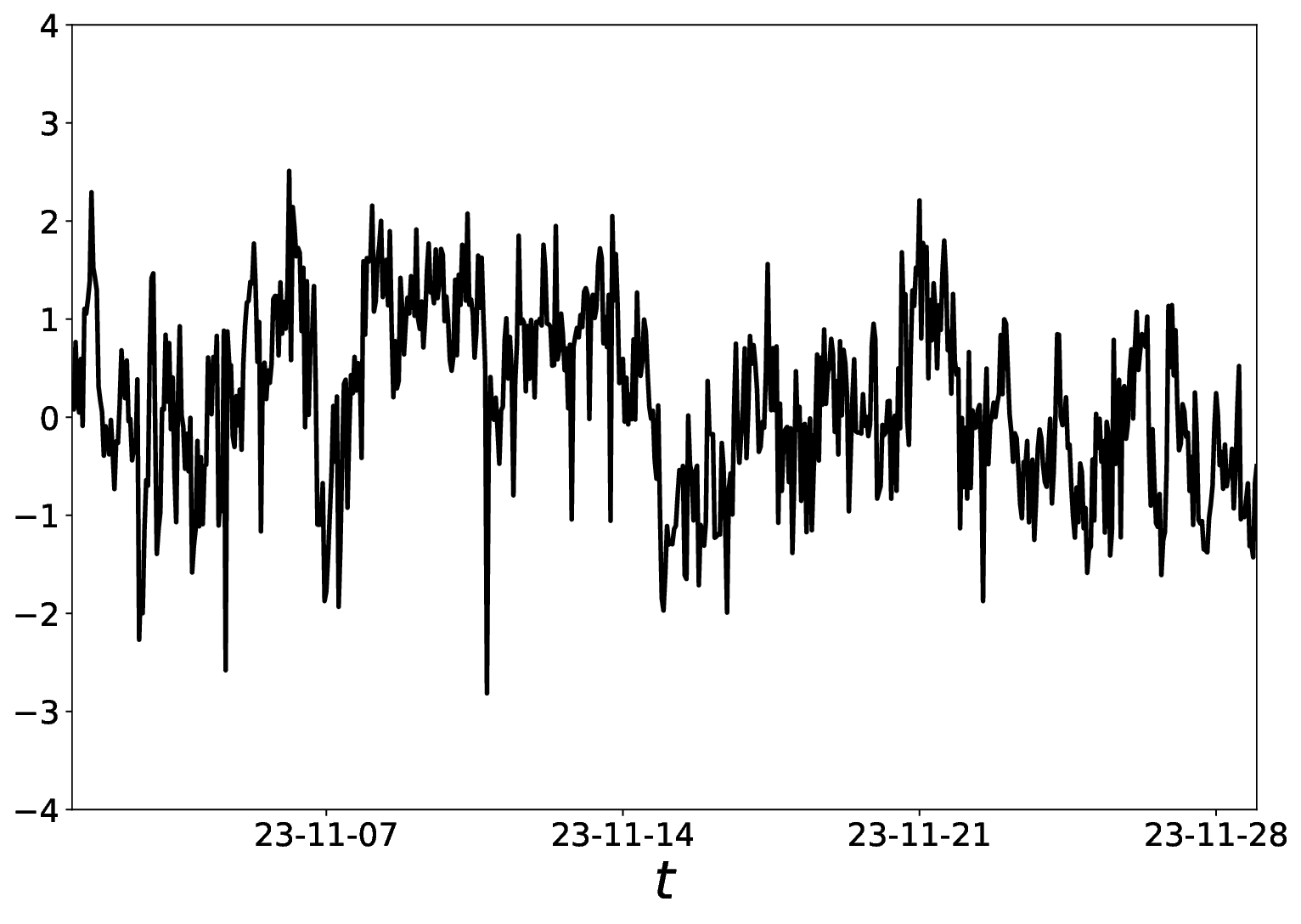}  
    \caption{Sample of $X_3(t)$}
\end{subfigure}
\begin{subfigure}{.31\textwidth}
    \centering
    \includegraphics[width=0.95\linewidth]{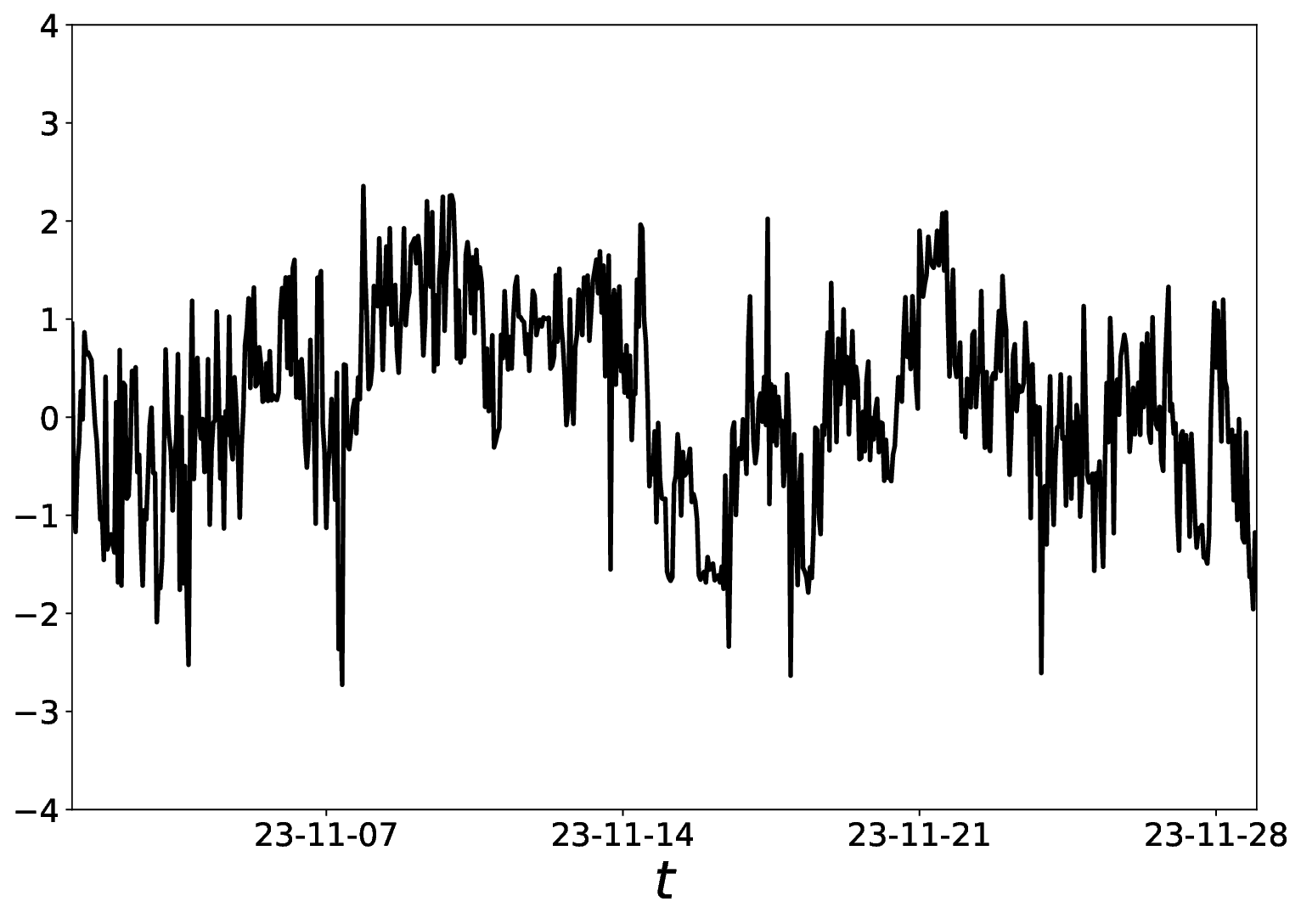}  
    \caption{Sample of $X_6(t)$}
\end{subfigure}
\caption{Synthetic realizations of $X_1(t), X_3(t), X_6(t)$ produced by the GenFormer model. The synthetic wind speed records appear realistic and can therefore be used for downstream applications of interest. }
\label{fig:data_samples_toy_wind}
\end{figure}

Figure~\ref{fig:data_samples_toy_wind} shows the records of the synthetic wind speed data in the Gaussian space produced by the GenFormer model at selected stations. A visual inspection of these time series, compared to the collected wind speed data shown in Figure~\ref{fig:ts_dl_toy_wind}, demonstrates that the synthetic samples appear realistic and look similar to the given observations. These synthetic samples can thus be used for various downstream applications. 

\begin{figure}[h]
\centering
\includegraphics[width=9cm]{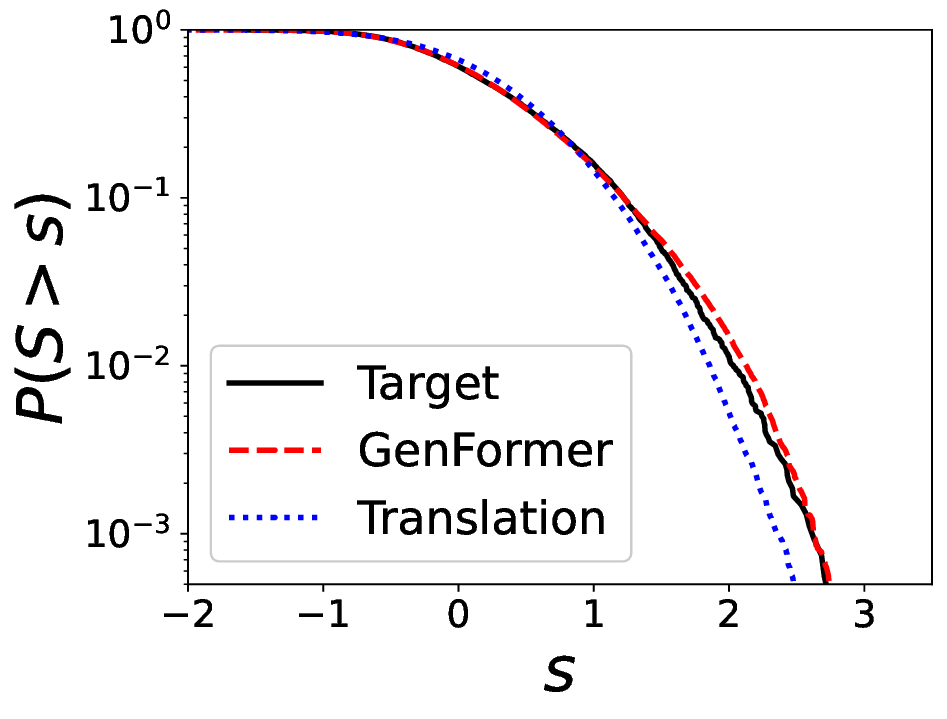}
\caption{Exceedance probability of $S$. The estimate obtained from the Genformer model is $7.6$ times more accurate than the translation model. The predictive capabilities of the deep learning model coupled with the statistical post-processing techniques enable the GenFormer model to capture high-order statistical properties while the translation model can only match up to the second-moment properties. }
\label{fig:metric_of_interest_toy_wind}
\end{figure}

The downstream application we pursue in this example is defined as the maximum of the monthly-averaged hourly wind speeds across stations given by $S = \max\limits_{1 \le i \le 6} \{ \int_0^{\timeEndSim} X_i(t) dt / \timeEndSim \}$. Such a metric can be used as a measure of the local hurricane intensity which is of interest in parametric insurance \cite{parametric_insurance}. In Figure~\ref{fig:metric_of_interest_toy_wind}, we plot the exceedance probabilities of the metric of interest $p(s) = P(S > s)$ computed using the observed data and synthetic realizations generated by the GenFormer model. The translation process is adopted as the baseline model for comparison. Similar to the previous example, we see that the approximation due to the translation model deteriorates as $s$ increases while the proposed GenFormer model is able to provide an accurate estimate. This is further evidenced by the return-period-based relative $L_1$ error of the translation process which is 0.9713 compared to that of the GenFormer which is 0.1281, a 7.6 times improvement over the former model. This improvement can become more substantial at specific values of $s$. For example, at $s = 2.7$ with a target return period of 1684 months, the estimates from the GenFormer and the translation model are 1376 and 6088 months, reflecting a 14-fold enhancement in the relative error. This improved accuracy is attributed to the capability of GenFormer in capturing higher-order statistical properties, even when $\nSites$ is large.

\section{Conclusion}
\label{sec:conclusion}

We presented GenFormer, a novel stochastic generator for producing synthetic realizations of spatio-temporal multivariate stochastic processes. The model integrates a univariate discrete-time Markov process capturing spatial variation with a Transformer-based deep learning model mapping the Markov states to time series values. GenFormer offers a scalable alternative for simulating multivariate processes in high dimensions and long horizons as well as Markov state sequences of large Markov orders. It exploits the predictive power of deep learning models coupled with modern computing capabilities while leveraging statistical post-processing techniques to guarantee that key statistical behavior is preserved. Numerical experiments applying the GenFormer model to simulate wind speeds in Florida demonstrated its ability to produce samples that approximate statistical properties beyond the second moment, unlike traditional methods. The GenFormer model can thus be reliably employed in various downstream applications in engineering. Utilizing state-of-the-art attention mechanisms in the Transformer architecture can further improve the performance of the proposed approach.

\bibliographystyle{plain}
\bibliography{main}

\end{document}